\documentclass{article}
\usepackage{preamble}

\title{Direct Evolutionary Optimization of\\{}Variational\,Autoencoders\,With\,Binary\,Latents}

\author{Enrico Guiraud \thanks{also affiliated with the Machine Learning Lab, University of Oldenburg, Germany.} \\
CERN, 1211 Geneva 23, Switzerland\\
\texttt{enrico.guiraud@cern.ch} \\
\And
Jakob Drefs \& J\"org L\"ucke\\
Machine Learning, University of Oldenburg, Germany \\
\texttt{\{jakob.drefs,joerg.luecke\}@uol.de} \\
}

\title{\vspace{-0pt}\LARGE \textbf{Direct Evolutionary Optimization of\\Variational\,Autoencoders\,With\,Binary\,Latents}}

\author{
\large
Enrico Guiraud$^{1,2}$, Jakob Drefs$^{1}$, J\"org L\"ucke$^{1}$ \\
\normalsize enrico.guiraud@cern.ch, joerg.luecke@uol.de, jakob.drefs@uni-oldenburg.de \\[2mm]
\normalsize $^1$\,Machine Learning Lab, University of Oldenburg, Germany \\
\normalsize $^2$\,CERN, Switzerland
\vspace{20pt}
}

\date{}

\begin{document}

\maketitle

\begin{abstract}
Discrete latent variables are considered important to model the generation process of real world data,
which has motivated research on Variational Autoencoders (VAEs) with discrete latents.
However, standard VAE training is not possible in this case, which has motivated different
strategies to manipulate discrete distributions in order to train discrete VAEs similarly to conventional ones.
Here we ask if it is also possible to keep the discrete nature of the latents fully intact by applying a direct
discrete optimization for the encoding model. The studied approach is consequently strongly diverting
from standard VAE training by altogether sidestepping absolute standard VAE mechanisms such as sampling approximation,
reparameterization trick and amortization.
Discrete optimization is realized in a variational setting using truncated posteriors in conjunction with evolutionary algorithms (using
a recently suggested approach). For VAEs with binary latents, we first show how such a discrete variational method (A)~ties into gradient
ascent for network weights and (B)~uses the decoder network to select latent states for training.
More conventional amortized training is, as may be expected, more efficient than direct discrete optimization, and applicable
to large neural networks. However, we here find direct optimization to be efficiently scalable to hundreds of latent variables using smaller networks.
More importantly, we find the effectiveness of direct optimization to be highly competitive in `zero-shot' learning (where high effectiveness
for small networks is required). In contrast to large supervised neural networks, the here investigated VAEs can, e.g., denoise a single image without
previous training on clean data and/or training on large image datasets.
More generally, the studied approach shows that training of VAEs is indeed possible without sampling-based approximation and reparameterization,
which may be interesting for the analysis of VAE-training in general. In the regime of few data, direct optimization, furthermore, 
makes VAEs competitive for denoising where they have previously been outperformed by non-generative approaches.
\blankfootnote{
  Preliminary version. Final version published as Drefs, J., Guiraud, E., Panagiotou, F., Lücke, J. (2023). Direct Evolutionary Optimization of Variational Autoencoders with Binary Latents. In: Amini, MR., Canu, S., Fischer, A., Guns, T., Kralj Novak, P., Tsoumakas, G. (eds) \emph{Machine Learning and Knowledge Discovery in Databases. ECML PKDD 2022. Lecture Notes in Computer Science}, vol 13715. Springer, Cham. \url{https://doi.org/10.1007/978-3-031-26409-2_22}
}
\end{abstract}

\section{Introduction and Related Work}
Variational autoencoders \citep[][]{KingmaWelling2014,Rezende2014} are prominent and very actively researched models for unsupervised learning. VAEs, in their
many different variations, have successfully been applied to a large number of tasks including semi-supervised learning \citep[e.g.][]{maaloe2016auxiliary}, anomaly detection \citep[e.g.][]{an2015variational,kiran2018overview}, sentence interpolation \citep{bowman2016generating}, music interpolation \citep{roberts2018hierarchical} and drug response prediction \citep{rampasek2017drvae}.
The success of VAEs rests on a series of methods that enable the derivation of scalable training algorithms to optimize their model parameters (discussed further below).
A desired feature when applying VAEs to a given problem is that their latent variables (i.e., the encoder output variables) correspond to meaningful properties
of the data, ideally to those latent causes that have originally generated the data.
However, many real-world datasets suggest the use of \emph{discrete} latents as they often describe the data generation process more naturally. For instance, the presence or absence of objects in images is best described by binary latents \citep[e.g.][]{JojicFrey2001}. Discrete latents are also a popular choice in modeling sounds; for instance, describing piano sounds may naturally involve binary latents: keys are pressed or not \citep[e.g.][]{TitsiasLazaro2011,GoodfellowEtAl2013,SheikhEtAl2014}.
The success of standard forms of VAEs has consequently spurred research on novel formulations that feature discrete latents \citep[e.g.][]{Rolfe2017,Khoshaman2018,RoyEtAl2018,Sadeghi2019,VahdatEtAl2019}.
%
%increasing body of research on variations of train procedures but
%also variants of VAE data models which include VAEs with discrete latents \citep[e.g.][]{...}. 
%
%Further examples, also beyond binary latents, may also be found in work directly related to VAEs \citep[][]{...}.
%
%As many causes of data are inherently discrete or binary,
%such as absence or presence of specific objects and object parts, VAEs with binary or discrete latents are of interest \citep[][]{}... .
%
%In addition to numerous variants of VAEs with continuous latents, the success 
%
%
%The success of VAEs with continuous latents (e.g., the standard VAE of Eqns.\,\ref{EqnDecodingVAEPrior} and \ref{EqnDecodingVAENoise}) has spurred an increasing body of research on new %variants of VAEs including an increased recent interest in VAEs with discrete latents \citep[e.g.][]{...}. 

The objective of VAE training is the optimization of a generative data model % (more precisely a {\em latent variable generative model})
which parameterizes a given data distribution. % using prior, noise model and deep neural networks (DNNs).
%; in standard VAEs, data noise is typically assumed Gaussian).
Typically we seek model parameters $\Theta$ of a VAE that maximize the data log-likelihood, $\textstyle\LL(\Theta) = \sum_{n} \log\big(\pTheta(\xVecN)\big)$, 
%
%\begin{equation}
%%
%%\LL(\Theta) = \sum_{n=1}^{N} \log\big(\pTheta(\xVecN)\big),   %\ =\ \sum_n \log\big(\\p(\yVecN\,|\,\Theta)
%\textstyle\phantom{xxxxx}\LL(\Theta) = \sum_{n} \log\big(\pTheta(\xVecN)\big),   %\ =\ \sum_n \log\big(\\p(\yVecN\,|\,\Theta)
%%
%\label{EqnLL}
%\end{equation}
%
where we denote by $\xVec^{(1:N)}$ a set of $N$ observed data points, and where $\pTheta(\xVec)$ denotes
the modeled data distribution. % provided by a given VAE.
Like conventional autoencoders \citep[e.g.,][]{bengio2007greedy}, VAEs use a deep neural network (DNN) to generate (or decode) observables $\xVec$ from a latent code $\vec{z}$.
Unlike conventional autoencoders, however, the generation of data $\xVec$ is not deterministic but it takes the form of a probabilistic generative model. 

For VAEs with binary latent variables, as they will be of interest here, we consider the following VAE generative model:\vspace{0mm}
\begin{equation}
    \pTheta(\zVec) = \Bern(\zVec;\piVec) = \textstyle\prod_{h} \big(\pi_h^{z_h} (1-\pi_h)^{(1 - z_h)}\big), 
   \,\,\,\,\,\,\,\,\,\,\,\,\,\,\,\,
    \pTheta(\xVec\,|\,\zVec) = \NCal\big(\xVec;\muVec(\zVec; W), \sigma^2 \mathbb{I} \big),
    \label{EqnBinaryVAE}
\end{equation}
%
%\begin{eqnarray}
%
%\zVec &\sim& \Bern(\zVec;\piVec) = \prod_{h=1}^H \pi_h^{z_h} (1-\pi_h)^{(1 - z_h)} \label{EqnEncodingBVAENoise}\\
%
%\xVec &\sim& \NCal(\xVec;\fVec(\zVec,W),\sigma^2\dblone), \label{EqnDecodingBVAEPrior}
%
%\end{eqnarray}
%
where $\zVec\in\{0,1\}^H$ is a binary code and the non-linear function $\muVec(\zVec; W)$ is a DNN that outputs the mean of the Gaussian distribution.
$\pTheta(\xVec\,|\,\zVec)$ is commonly referred to as {\em decoder}.
The set of model parameters is $\Theta=\{\piVec, W, \sigma^2\}$, where $W$ incorporates DNN weights and biases. We assume homoscedasticity of the Gaussian distribution, but note that there is no obstacle to generalizing the model by inserting a DNN non-linearity that outputs a correlation matrix. 
%One advantage of employing a simple scalar variance is that analytical update rules can be derived in closed form.
Similarly, the algorithm could easily be generalized to different noise distributions should the task at hand call for it. For the purpose of this work, however, we will focus on as elementary as possible VAEs, with the form shown in Eqn.\, \refp{EqnBinaryVAE}.
%
%The log-likelihood of the VAE data model is given by
%(\ref{EqnLL}) where $\pTheta(\xVec) = \sum_{\zVec}\pTheta(\xVec\,|\,\zVec)\,\pTheta(\zVec)$ ($\sum_{\zVec}$ sums over all possible binary vectors).
%
%\begin{eqnarray}
%
%\pTheta(\xVec) &=& \sum_{\zVec}\pTheta(\xVec\,|\,\zVec)\,\pTheta(\zVec),
%
%\label{EqnModelDis}
%\end{eqnarray}
%where $\sum_{\zVec}$ sums across all possible binary states. 

Given standard or binary-latent VAEs, essentially all learning algorithms seek to approximately maximize the log-likelihood %(\ref{EqnLL})
using the following series of methods (we elaborate in the appendix):\vspace{-1mm}
\begin{itemize}
\item[(A)] Instead of the log-likelihood, a variational lower-bound (a.k.a.\ ELBO) is optimized.\vspace{-1mm}
\item[(B)] VAE posteriors are approximated by an {\em encoding model}, that is a specific distribution (often Gaussian) parameterized by one or more DNNs. \vspace{-1mm}
%
%   distributions (often a Gaussian) are assumed as variational approximations to the posteriors defined by the VAE's decoding model.
%	   These distributions are            
%           The assumed variational distributions contain non-linearities defined by DNNs (usually for mean and variance), which make up the
%           {\em encoding model} of the VAE.
%
%An {\em amortized} \citep[][]{...} function ()
%As encoding model is {\em amortized} \citep[][]{...}, i.e., {\em one} parametric encoding model applies for data points $\yVecN$
%	   instead of using one sets of variational parameters for each data point (as done, e.g., for conventional mean field or Gaussian approaches \citep[][]{}).
%
%\item[(D)] The variational lower bound is optimized w.r.t.\ the variational distribution and the paramters of the decoding model.
%alternately w.r.t.\ the variational distribution and the paramters of the decoding model.
%
%\item[(D)] The model parameters of the lower bound are optimized (in alteration with the variational parameters) using gradient ascent.
%
\item[(C)] The variational parameters of the encoder are optimized using gradient ascent on the lower bound, where
the gradient is evaluated based on sampling and reparameterization trick to obtain sufficiently low-variance and yet efficiently computable estimates. \vspace{-1mm}
%
%, which is made possible by the reparameterization trick \citep[][]{KingmaWelling2014,Rezende2014}, which in turn enables a sampling approximation of the lower bound that yields sufficiently low-variance gradients\vspace{-1mm}
%
\item[(D)] Using samples from the encoder, the parameters of the decoder are optimized using gradient ascent on the variational lower bound.\vspace{-2mm}
%Notably, variational and model parameters can be optimized jointly, via one application of the gradient operator.
%
\end{itemize}
%
%Alternatively to the reparameterization trick, the log-derivative trick and variance reduction techniques to reduce the large variance of the estimated lower bound gradients can be used \citep[][]{Williams1992}.
%
Optimization procedures for VAEs with discrete latents follow the same steps (Points A to D).
However, discrete or binary latents pose substantial further obstacles in learning, mainly due to the fact that backpropagation through discrete variables is generally not possible \citep{Rolfe2017,bengio2013estimating}.
In order to maintain the general VAE framework for encoder optimization, different groups have therefore suggested different possible solutions: work by \citet[][]{Rolfe2017}, for instance, extends VAEs with discrete latents by auxiliary continuous latents such that gradients can still be computed. Work on the concrete distribution \citep[][]{Maddison2017} or Gumbel-softmax distribution \citep[][]{Jang2017} proposes newly defined continuous distributions that contain discrete distributions as limit cases. Work by \cite{lorberbom2019direct} merges the Gumbel-Max reparameterization with the use of direct loss minimization for gradient estimation, enabling efficient training on structured latent spaces. Finally, work by \citet{OordEtAl2017}, and \citet{RoyEtAl2018} combines VAEs with a vector quantization (VQ) stage in the latent layer. Latents become discrete through quantization but gradients for learning are adapted from latent values before they are processed by the VQ stage.
All methods have the goal of treating discrete distributions such that standard VAE training as developed for continuous latents can still be applied.
%The continuous distributions can then be treated within the standard VAE framework. 
%
%All these approaches maintain the basic setup of VAEs but add specific additional mechanisms to treat discrete latents.
These techniques interact during training with the standard methods (Points A-D) already in place for VAE optimization. % (Point D).
Furthermore, they add further types of design decisions and hyper-parameters, for example parameters for annealing from softened discrete distributions to the (hard) original distributions for discrete latents.

For discrete VAEs, it may consequently be a desirable goal to investigate alternative, more direct optimization procedures that do not require
a softening of discrete distributions or the use of other indirect solutions.
Such a direct approach is challenging, however, because once DNNs are used to define
the encoding model (Point~B) standard tricks to estimate gradients (Point~C) seem unavoidable.
A direct optimization procedure, as is investigated here, consequently has to substantially change VAE training. For the data model (\ref{EqnBinaryVAE}) we will maintain the variational setting and a decoding model with DNNs as non-linearity (Points~A and D).
However, we will not use an encoder model parameterized by DNNs (Point~B). Instead, the variational bound will be increased w.r.t.\ the encoder model by using a discrete optimization approach.
The procedure does not require gradients to be computed for the encoder such that discrete latents are addressed without the use of reparameterization trick and sampling approximations.

\section{Truncated Variational Optimization}
%
%efficient are variational distributions in the form of
%truncated posteriors. Truncated posteriors have previously been applied to optimize generative models with discrete latents \citep[e.g.][]{SheikhEtAl2014,HughesSudderth2016,LuckeEtAl2017,SheltonEtAl2017,ForsterLucke2018}. 
%
%The key ingredient that enables a discrete optimization of the encoder is a specific
%choice of encoding model \qPhi$. 
%
%, namely truncated posteriors.
%We refer to the generative model (\ref{EqnBinaryVAE}), trained with truncated posteriors and the discrete
%optimization algorithm that they unlock, as Truncated Variational Autoencoders (or {\em TVAE} for short).

%\subsection{Direct Optimization of the Encoding Model}
%
%n this work we focus on VAEs with binary latents and follow an alternative approach for the optimization of the encoding model. We maintain the decoding model of VAEs (Fig. X) with
%Bernoulli distributions as prior for the binary latents, and a Gaussian distribution with a DNN non-linearity for the data. Also the approach to optimize the parameters $\Theta$ is
%similar in its form to standard VAEs: we compute the gradient of the lower bound $\FF(\Phi,\Theta)$ w.r.t.\,$\Theta$, which translates to a gradient w.r.t.\,the used DNN
%(we will elaborate further below).
%
%plus gradients w.r.t.\,other parameters such
%as noise variance).
%
%However, instead of aiming to optimize the encoding model $\qPhi(\zVec;\xVec)$ using the series of reformulation steps and tools listed in Fig.\,\ref{FigEncoding}A,B, we approach the optimization of VAEs with binary latents more directly. 
%
Let us consider the variational lower bound of the likelihood. If we denote by $\qPhiN(\zVec)$
the variational distributions with parameters $\PhiN$, and by $\EE{ h(\zVec) }_{\qPhiN}\,=\,\sum_{\zVec}\ \qPhiN(\zVec)\ h(\zVec)$ expectation values w.r.t.\ to $\qPhiN(\zVec)$, then the lower bound can be written as:
\begin{equation} 
%
%  \FF(\Phi,\Theta) &=& \sum_{n=1}^{N} \sum_{\zVec}\ \qPhiN(\zVec)\ \log\left( \pTheta(\xVecN\,|\,\zVec) \, \pTheta(\zVec) \right)
%                    - \sum_{n=1}^{N} \sum_{\zVec}\ \qPhiN(\zVec)\ \log\left( \qPhiN(\zVec) \right) \nonumber\\
%
\textstyle\FF(\Phi,\Theta) = \textstyle\sum_{n} \EE{ \log\big( \pTheta(\xVecN\,|\,\zVec)\,\pTheta(\zVec) \big) }_{\qPhiN}
\textstyle                   \textstyle-\, \sum_{n} \EE{ \log\big( \qPhiN(\zVec) \big) }_{\qPhiN},
                   \label{EqnFWithExp}
%
%		&& \mbox{where\ \ \ } \EE{ h(\zVec) }_{\qPhiN}\,=\,\sum_{\zVec}\ \qPhiN(\zVec)\ h(\zVec)
%
\end{equation}
The general challenge for the maximization of $\FF(\Phi,\Theta)$ is the optimization of the encoding model $\qPhiN$. VAEs with discrete latents
add to this challenge the problem of taking gradients w.r.t.\ discrete latents. If we seek to avoid derivatives w.r.t.\ discrete variables,
we have to define an alternative encoding model $\qPhiN$ but such an encoding has to remain sufficiently efficient. Considering prior work on
generative models with discrete latents, variational distributions based on truncated posteriors offer themselves as such an alternative \citep[][]{LuckeSahani2008}.
Truncated posterior approximations have been shown to be functionally competitive \citep[e.g.][]{SheikhEtAl2014,HughesSudderth2016,SheltonEtAl2017}, and they are able
to efficiently train also very large scale models with hundreds or thousands of latents \citep[e.g.][]{SheltonEtAl2011,SheikhLucke2016,ForsterLucke2018}.
However, the important question for training discrete VAEs is if or how truncated variational distributions can be used in gradient-based optimization of neural network parameters. We here, for the first time, address this question noting that all previous approaches relied on closed-form (or pseudo-closed form) parameter update equations in an expectation-maximization learning paradigm.

{\bf Optimization of the Decoding Model.} In order to optimize the parameters $W$ of the decoder DNN $\vec{\mu}(\zVec,W)$, the gradient of the variational bound (\ref{EqnFWithExp}) w.r.t.\ $W$ has to be computed.
We consequently need, for any VAE, a sufficiently precise and efficient approximation of the expectation value w.r.t.\ the encoder $\qPhiN(\zVec)$.
Gradient estimation is of central importance for deep unsupervised learning, and approaches, e.g., for variance reduction of estimators have played
an important role and are dedicated solely to this purpose \citep[e.g.,][]{Williams1992}. Reparameterization finally emerged as a key method because
it allowed for
sufficiently low-variance estimation of gradients based, e.g., on Gaussian middle-layer units \citep[][]{KingmaWelling2014,Rezende2014}.
%Similar ideas have been used before \citep[][]{PapaspiliopoulosEtAl2003,ChallisBarber2012} but the achievement of \citet[][]{KingmaWelling2014,Rezende2014} was to show how a deep generative model can be trained using reparameterization.

For discrete VAEs, however, reparameterization requires the introduction of additional manipulations of discrete distributions that we here seek to fully avoid.

Instead of using reparameterization or variance reduction, we will compute gradients based on truncated posterior as variational distributions. A truncated posterior
has the following form:
\begin{equation}
\qPhiN(\zVec) := \phantom{\int} \frac{\pTheta(\zVec\,|\,\xVecN)}{\sum_{\zVecPrime\in\PhiN}\,\pTheta(\zVecPrime\,|\,\xVecN)}
\,=\,\frac{\pTheta(\xVecN\,|\,\zVec)\,\pTheta(\zVec)}{\sum_{\zVecPrime\in\PhiN}\,\pTheta(\xVecN\,|\,\zVecPrime)\,\pTheta(\zVecPrime)}    \,\ \mbox{\ if\ }\  \zVec\in\PhiN,
%
%\qPhiN(\zVec) := \phantom{\int}\pTheta(\zVec\,|\,\xVecN)\,\,\,\,\, \Big/ \,\, \sum_{\zVecPrime\in\PhiN}\,\pTheta(\zVecPrime\,|\,\xVecN)\,\ \mbox{\ if\ }\  \zVec\in\PhiN,
%                   &=\frac{\disS\phantom{\int}\pTheta(\xVecN\,|\,\zVec)\ \pTheta(\zVec)\phantom{\int}}{\hspace{-2mm}\disS\sum_{\zVecPrime\in\PhiN}\pTheta(\xVecN\,|\,\zVecPrime)\ \pTheta(\zVecPrime)}\,\delta(\zVec\in\PhiN),
%
\label{EqnQMain}
\end{equation}
where for all $\zVec\not\in\PhiN$ the probability $\qPhiN(\zVec)$ equals zero. That is, a variational distribution $\qPhiN(\zVec)$ is proportional to
the true posteriors in a subset $\PhiN$, which acts as its variational parameter.

%The expectation of the log-joint is then given by:
%
%
%\begin{equation}
%    \left<h(\zVec\,)\right>_{q_\Phi} = \frac{\sum\limits_{\zVec \in \Phi} h(\zVec\,) p_\Theta(\zVec, \xVec)}{\sum\limits_{\zVec^{\,\prime} \in \Phi} p_\Theta(\zVec^{\,'}, \xVec)} 
%\EE{ \log\big( \pTheta(\xVecN\,|\,\zVec)\,\pTheta(\zVec) \big) }_{\qPhiN}\ =\ \sum\limits_{\zVec \in \PhiN} \qPhiN(\zVec)\phantom{i} \log\big( \pTheta(\xVecN\,|\,\zVec)\,\pTheta(\zVec) \big)
%    \label{EqnMeanPosterior}
%\end{equation}
%
We can now compute the gradient of (\ref{EqnFWithExp}) w.r.t.\ the decoder weights $W$ which results in:
\begin{equation}
  \vec{\nabla}_{W}\FF(\Phi,\Theta) %= \sum_{n}  \EE{ \vec{\nabla}_{W}\log\big( \NCal(\xVecN;\vec{\mu}(\zVec,W),\sigma^2\mathbb{I} \big) }_{\qPhiN} \nonumber\\
                                   = -\frac{1}{2\sigma^2} \sum_{n} \sum\limits_{\ \zVec \in \PhiN} \qPhiN(\zVec)\phantom{i}\vec{\nabla}_{W} \|\xVecN - \vec{\mu}(\zVec,W)\|^2\,. \label{EqnFDeriW}
%
%                                      &=& \phantom{+}\sum_{n}\frac{\sum_{\zVec\in\PhiN} \pTheta(\xVecN\,|\,\zVec)\,\pTheta(\zVec) \ \vec{\nabla}_{\Theta} \log\big( \pTheta(\xVecN\,|\,\zVec) \big)}
%                                                                  {\sum_{\zVec\in\PhiN} \pTheta(\xVecN\,|\,\zVec)\,\pTheta(\zVec)} \\
%
%                                       &&          + \sum_{n}\frac{\sum_{\zVec\in\PhiN}\pTheta(\xVecN\,|\,\zVec)\,\pTheta(\zVec) \ \vec{\nabla}_{\Theta} \log\big( \pTheta(\zVec) \big)}
%                                                                  {\sum_{\zVec\in\PhiN}\pTheta(\xVecN\,|\,\zVec)\,\pTheta(\zVec)}
%
\end{equation}
The right-hand-side has salient similarities to the standard gradient ascent for VAE decoders. Especially the familiar gradient of the mean squared error (MSE) shows that, e.g., standard automatic differentiation tools can be applied. However, the decisive difference are the weighting factors $\qPhiN(\zVec)$. Considering (\ref{EqnQMain}), in order to compute the weighting factors we require all $\zVec\in\PhiN$ to be passed through the decoder DNN. As all states of $\PhiN$ anyway
have to be passed through the decoder for the MSE term of (\ref{EqnFDeriW}), the overall computational complexity is not higher than an estimation of the gradient
with samples instead of states in $\PhiN$ (we elaborate in Appendix A).

To complete the decoder optimization, update equations for variance $\sigma^2$ and prior parameters $\vec{\pi}$ can be computed in closed-form \citep[compare, e.g.,][]{SheltonEtAl2011} and are given by:
\begin{equation}
  \sigma^{2,\mathrm{new}} = \textstyle \frac{1}{DN} \sum\limits_{n} \sum\limits_{\ \zVec \in \PhiN} \qPhiN(\zVec)\   \| \xVecN - \vec{\mu}(\zVec,W)\|^2\,,
  \,\,\,\,\,\,\, \piVec^{\mathrm{new}} = \textstyle \frac{1}{N} \sum\limits_{n} \sum\limits_{\ \zVec \in \PhiN} \qPhiN(\zVec)\ \zVec,
\label{EqnPiSigma}
\end{equation}
where $N$ is the number of samples in the training dataset and $D$ is the number of
observables.% (the dimensionality of $\xVec$).

% While $\sigma^2$ and $\piVec$ can (like $W$)
%be updated with every batch, they are here for simplicity updated at the end of every epoch. 

{\bf Optimization of the Encoding Model.} After having established that the decoder can be optimized efficiently and by using standard DNN methods,
the important question is if the encoder can be trained efficiently. Encoder optimization is usually based on a reformulation of the variational bound (\ref{EqnFWithExp}) given by: % \citep[][]{KingmaWelling2014,Rezende2014}:
\begin{equation}
\textstyle  \FF(\Phi,\Theta) =\textstyle \sum_{n} \EE{\log\big( \pTheta(\xVecN\,|\,\zVec) \big)}_{\qPhiN}
\textstyle                    \textstyle- \sum_{n} \DKL{\qPhiN(\zVec)}{\pTheta(\zVec)}.
  \label{EqnFVAE}
\end{equation}
Centrally for this work, truncated posteriors allow a specific alternative reformulation of the bound that enables efficient optimization.
The reformulation recombines the entropy term of the original form (\ref{EqnFWithExp})
with the first expectation value into a single term, and is given by \citep[see][for details]{Lucke2019}: 
\begin{equation}
\label{EqnFMain}
\FF(\Phi,\Theta) = \disS\sum_{n}\log\big(\backHalf\sum_{\ \zVec\in\PhiN} \pTheta(\xVecN\,|\,\zVec)\,\pTheta(\zVec)\big)\,.
\end{equation}
%
%
%Unlike the original form of the lower bound (\ref{EqnFWithExp}) and unlike the VAE form (\ref{EqnFVAE}), the reformulation of Eqn.\,\ref{EqnFMain} now contains just a single term.
%While the truncated encoding distributions (\ref{EqnQMain}) do have non-zero entropy, the entropy term of the original form (\ref{EqnFWithExp})
%can be recombined with the first expectation value \citep[see][for details]{Lucke2019}.
%
Thanks to the simplified form of the bound, the variational parameters $\PhiN$ of the encoding model can now be sought using direct discrete optimization procedures. More concretely, because of the specific form (\ref{EqnFMain}), pairwise comparisons of joint probabilities are
sufficient to maximize the lower bound: if we update the set $\PhiN$ for a given $\xVecN$ by replacing a state $\zVecOld\in\PhiN$ with a state $\zVecNew\not\in\PhiN$,
then $\FF(\Phi,\Theta)$ increases if and only if:
\begin{equation}
%
%  \pTheta(\xVecN\,|\,\zVecNew)\,\pTheta(\zVecNew) &>& \pTheta(\xVecN\,|\,\zVecOld)\,\pTheta(\zVecOld)\,.\label{EqnCriterion}\\
%
%  \pTheta(\xVecN, \zVecNew) > \pTheta(\xVecN, \zVecOld)\\
%  \text{or, equivalently,}\\
\log\big( \pTheta(\xVecN, \zVecNew) \big) > \log\big( \pTheta(\xVecN, \zVecOld) \big) \label{EqnCriterion}\,.
%
%\ \ \Leftrightarrow\ \ \qPhiN(\zVecNew) > \qPhiN(\zVecOld) \label{EqnCriterion}\,.\\
%
%
\end{equation}
To obtain intuition for the pairwise comparison, consider its form when inserting the binary VAE \refp{EqnBinaryVAE} into the left- and right-hand sides. Eliding terms that do not depend on $\zVec$ we obtain:
\begin{equation}
%
%  \log\big(\pTheta(\xVecN\,|\,\zVecNew)\big) + \log\big(\pTheta(\zVecNew)\big) &>&   \log\big(\pTheta(\xVecN\,|\,\zVecOld)\big) + \log\big(\pTheta(\zVecOld)\big)\\
% &\|\xVecN-\vec{\mu}(\zVecNew,W)\|^2\,+\,2\,\sigma^2 \sum_{h=1}^H \tilde{\pi}_h\,\zNew_h \, &<& \, &\|\xVecN-\vec{\mu}(\zVecOld,W)\|^2\,+\,2\,\sigma^2 \sum_{h=1}^H \tilde{\pi}_h\,\zOld_h\,,
\textstyle\widetilde{\log p}_\Theta(\xVec, \zVec) = - \|\xVec-\vec{\mu}(\zVec,W)\|^2\,-\,2\,\sigma^2 \sum_{h} \tilde{\pi}_h\,z_h
\label{EqnCriterionVAE}
\end{equation}
where $\tilde{\pi}_h=\log\big( (1-\pi_h)/\pi_h \big)$. The expression assumes an even more familiar form if we restrict ourselves for a moment to sparse priors $\pi<\frac{1}{2}$, i.e., $\tilde{\pi}>0$. Criterion (\ref{EqnCriterion}) then becomes:
\begin{equation}
  \|\xVecN-\vec{\mu}(\zVecNew,W)\|^2\,+\,2\,\sigma^2 \tilde{\pi}{}\ |\zVecNew| \ \ <\ \ \|\xVecN-\vec{\mu}(\zVecOld,W)\|^2\,+\,2\,\sigma^2 \tilde{\pi}{}\ |\zVecOld|
  \label{EqnCriterionSimple}
\end{equation}
where $|\zVec|= \sum_{h=1}^H z_h$.
Such functions are routinely encountered in sparse coding or compressive sensing \citep[][]{EldarKutyniok2012}: for each set $\PhiN$ we seek those states $\zVec$ that are reconstructing $\xVecN$ well while 
being sparse ($\zVec$ with few non-zero bits). For VAEs, $\vec{\mu}(\zVecNew,W)$ is a DNN and as such much more flexible in matching the distribution
of observables $\xVec$ than can be expected from linear mappings. Furthermore, criteria like (\ref{EqnCriterionSimple}) usually emerge for maximum a-posteriori (MAP) training in sparse coding \citep[][]{OlshausenField1996}. In contrast, we here seek a \emph{population} of states $\zVec$ in $\PhiN$ for each data point.
%Like a population of sampled states, the sets $\PhiN$ thus allow for capturing a rich posterior structure. 
It is a consequence of the reformulated lower bound (\ref{EqnFMain}) that it remains optimal to evaluate joint probabilities (as for MAP) although the constructed  population of states $\PhiN$ can capture (unlike MAP training) a rich posterior structure.

%For our numerical experiments, we will use the general form (\ref{EqnCriterionVAE}). For the implementation, we have to pass novel states $\zVecNew$ through
%the decoder and if they compare favorally with old states in $\PhiN$, the set $\PhiN$ can be updated
%and increases the lower bound (\ref{EqnFMain}). Again (like for the decoder optimization) such pairwise
%comparisons (which can be done in bunches), are efficient for sufficiently small $\PhiN$.

%{\bf Evolutionary Optimization.} 
But how can new states $\zVecNew$ that optimize $\PhiN$ be found efficiently in high-dimensional latent spaces?
Random search and search by sampling has recently been explored for elementary generative models \citep[][]{LuckeEtAl2018}.
Here we will follow another recent suggestion \citep[][]{GuiraudEtAl2018} and make use of a search based on evolutionary algorithms (EAs). % to generate candidate states $\zVecNew$. 
In this setting we interpret sets $\PhiN$ as populations of binary genomes $\zVec$ and base the fitness function on Eqn.\,\refp{EqnCriterionVAE}.

%\begin{align} 
%\begin{split}
%\fit(\zVec; \Theta, \xVec) = &- \|\xVec - \vec{\mu}(\zVec,W)\|^2\\
%                             &- 2 \sigma^2 \sum_{h=1}^H z_h \log\left(\frac{1-\pi_h}{\pi_h}\right).
%\label{EqnFitness}
%\end{split}
%\end{align}
%

\begin{figure}[th]
\begin{minipage}{0.5\textwidth}
    \centering
    \def\svgwidth{\textwidth}
    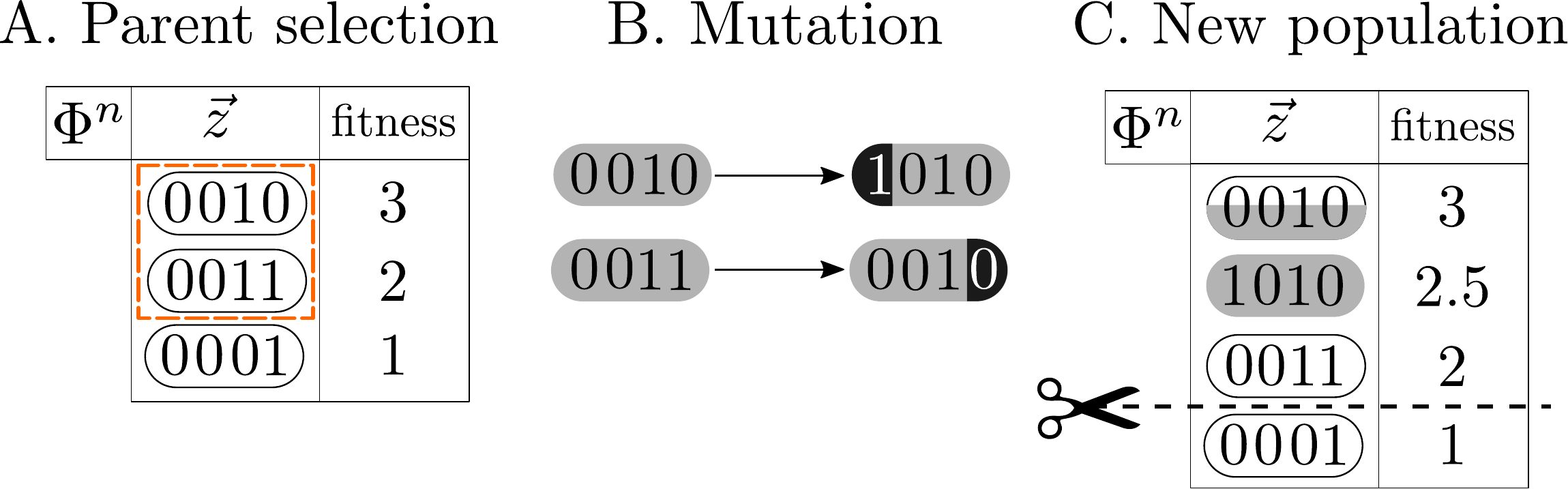
   \vskip -0.4cm
\hfill
\end{minipage}
\begin{minipage}{0.45\textwidth}
    \caption{The optimization process of the variational parameters $\Phi^{(n)}$ using evolutionary search. \textbf{A.} Some states are selected as parents. \textbf{B.} Each child undergoes mutation. \textbf{C.} Children are merged with the original population and the least fit are discarded.}
   \vskip -0.4cm
    \label{FigEEM}
\end{minipage}
\end{figure}

Concretely, using $\PhiN$ as initial parent pool, we apply the following genetic operators in sequence:
firstly, {\em parent selection} picks $N_p$ states from the parent pool. In our numerical experiments we used fitness-proportional parent selection, for which we add an offset (constant w.r.t.\ $\zVec$) to the fitness values in order to make them strictly non-negative.
Each of the children undergoes {\em mutation}: one or more bits are flipped to further increase offspring diversity. In our experiments we perform random uniform selection of the bits to flip.
Crossover could also be employed to increase offspring diversity.
We repeat the procedure using the children generated this way as the parent pool, giving birth to multiple \textit{generations} of candidate states. Finally, we update $\PhiN$ by substituting individuals with low fitness with candidates with higher fitness. The whole procedure can be seen as an evolutionary
algorithm with perfect memory or very strong elitism (individuals with higher fitness never drop out of the gene pool). Note that the improvement of the variational lower bound depends on generating as many as possible \textit{different} children with high fitness over the course of training.

We point out that the EAs optimize each $\PhiN$ independently, so this technique can be applied to large datasets in conjunction with stochastic or batch gradient descent on the model parameters $\Theta$: it does not require to keep the full dataset (or all sets $\PhiN$) in memory at a given time. Fig\,\ref{FigEEM} shows how EAs produce new states that are used to update each set $\PhiN$. The full training procedure for binary VAEs is summarized in Algorithm \ref{AlgEEM}.

%TODO: shorten paragraph below, move to intro or discussion
%Furthermore, states in $\PhiN$ can be combined with new states not in $\PhiN$ in bunches, and the $\PhiN$ can be updated by applying efficient selection algorithms \citep[][]{Blum1973}
%which scale linearly with the size of $\PhiN$. Recent such approaches thus gave rise to efficiently scalable algorithms for standard Bayesian networks
%such as noisy-OR or sigmoid belief networks \citep[][]{}, for clustering \citep[][]{ForsterLucke2018} as well as for discrete forms of sparse coding \citep[][]{SheikhEtAl2014,SheltonEtAl2017,LuckeEtAl2018}. If sets $\PhiN$ are coupled to a fast feedforward search for $\PhiN$ in the form of weighted
%sums \citep[][]{SheltonEtAl2011}, the largest sparse coding models to date could be trained \citep[][]{SheikhLucke2016}. Here we make use of the recent optimization
%of sets $\PhiN$ in \citep[][]{GuiraudEtAl2018} in order to find sets $\PhiN$. More concretely, we use ...

\hspace{-0.5cm}
\begin{minipage}{.60\textwidth}
\begin{algorithm}[H]
\caption{Training Truncated Variational Autoencoders}
\label{AlgEEM}
\begin{algorithmic} \small
\STATE Initialize model parameters $\Theta = \{ W, \piVec, \sigma^2 \}$
\STATE Initialize each $\PhiN$ with $S$ distinct latent states
\REPEAT
  \FORALL{batches in dataset}
    \FOR{sample $n$ in batch}
      \STATE $\Phi^{new} = \PhiN$
      \FORALL{generations}
        \STATE $\Phi^{new} = \text{mutation}\left(\text{crossover}\left(\text{selection} \left(\Phi^{new}\right)\right)\right)$\\
        \STATE $\PhiN\,=\,\PhiN \cup \Phi^{new}$
      \ENDFOR
      \STATE Truncate $\PhiN$ to $S$ fittest elements based on (\ref{EqnCriterionVAE})
    \ENDFOR
  \STATE Use Adam to update $W$ using objective (\ref{EqnFDeriW})
  \ENDFOR
\STATE Use (\ref{EqnPiSigma}) to update $\piVec$, $\sigma^2$
\UNTIL{parameters $\Theta$ have sufficiently converged}
\end{algorithmic}
\end{algorithm}
\end{minipage}
\hfill
\begin{minipage}{.40\textwidth}
\begin{figure}[H]
    \centering
    \def\svgwidth{\textwidth}
    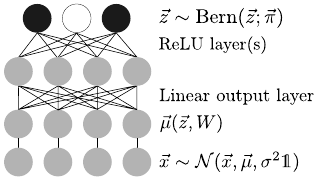
    \caption{Graphical representation of the model architecture
             used in numerical experiments.}
    \label{FigTVAE}
\end{figure}
\end{minipage}

\section{Numerical Experiments}
Having defined the training procedure, we numerically investigated its properties. After first verifying that the procedure can recover generating parameters using ground-truth data (see Appendix~B), we conducted experiments to address the following two standard questions:

(1) How efficient, i.e.\ how scalable, is the direct discrete optimization of binary VAEs?\\
(2) How effective is the procedure, i.e., how well does it perform for a given VAE model? 

In all numerical experiments, the training of the DNN parameters based on (\ref{EqnFDeriW}) is performed with mini-batches, the Adam optimizer \citep{kingma2014adam} and decaying or cyclical learning rate scheduling \citep{smith2017cyclical}. Xavier/Glorot initialization \citep{glorot2010understanding} is used for the DNN weights,
while biases are always zero-initialized. Parameters $\piVec$ and $\sigma^2$ are updated via Eqn.\,(\ref{EqnPiSigma}) and initialized to $\frac{1}{H}$ ($H$ is the size of $\vec{\pi}$) and $0.01$ respectively. Hyper-parameter optimization was conducted manually and, for the more complex datasets, it also made use of black box Bayesian optimization based on Gaussian Processes \citep{BayesOptGithub}.
%
%Training of the DNN using (\ref{EqnFDeriW}) is performed with mini-batches, the Adam optimizer \citep{kingma2014adam} and decaying or cyclical learning %rate scheduling \citep{smith2017cyclical}. Xavier/Glorot initialization \citep{glorot2010understanding} is used for the DNN weights,
%while biases are always zero-initialized. Parameters $\piVec$ and $\sigma^2$ are updated via Eqn.\,(\ref{EqnPiSigma}) and initialized to $\frac{1}{H}$ ($H$ is the size of $\vec{\pi}$) and $0.01$ respectively. Hyper-parameter optimization was conducted manually and, for the more complex datasets, it also made use of black box Bayesian optimization based on Gaussian Processes \citep{BayesOptGithub}.
%
We will refer to the binary VAE trained with the method described above as {\em Truncated Variational Autoencoder} (TVAE) as the use of truncated posteriors is the main distinguishing feature.

{\bf Scalability and improvement on linear models.} Let us first numerically investigate scalability properties of TVAE especially in comparison with
linear models. After verifying parameter recovery for ground-truth data (see Appendix B), we used natural data in the form of image patches as an intermediately large scale and natural dataset. %The simplest mapping $\vec{\mu}(\zVec,W)$ one can use is a linear mapping. 
Concretely, we used 100,000 whitened image patches of $16\times{}16$ pixels extracted from a standard image database \citep{HaterenSchaaf1998} and pre-processed as in \citet{GuiraudEtAl2018}. 

The simplest possible VAEs would use linear mappings for the decoder $\vec{\mu}(\zVec,W)$. For standard Gaussian latents, a linear VAE can recover
probabilistic PCA solutions \citep[e.g.][]{LucasEtAl2019}. For Bernoulli latents, we would recover binary sparse coding \citep[][]{HaftEtAl2004,SheltonEtAl2011} solutions. We therefore start training (using $H=300$ latents) with a linear VAE.
After 100 epochs the weights of the linear mapping were used to initialize
the bottom layer of a deeper decoder network with three layers of 300, 300 and $16\times{}16 = 256$ units. The weights of the deeper layers were simply initialized to the identity matrix. Furthermore, prior and variance were optimized. The described setup guarantees a common starting point for linear and non-linear VAEs such that the difference provided by deeper decoder DNNs can be highlighted. Fig.\,\ref{FigNaturalImages} shows the variational
bounds during learning of the linear VAE compared to the non-linear VAE for a typical experiment. The non-linear VAE can be observed to quickly and significantly optimize the lower bound beyond a linear VAE. We will later (when we are not interested in comparisons to linear VAEs) simply
optimize the weights of non-linear TVAE directly as we did not observed an advantage by first optimizing a linear VAE.

\begin{minipage}{0.45\linewidth}
\begin{figure}[H]
    \centering
    \def\svgwidth{\textwidth}
    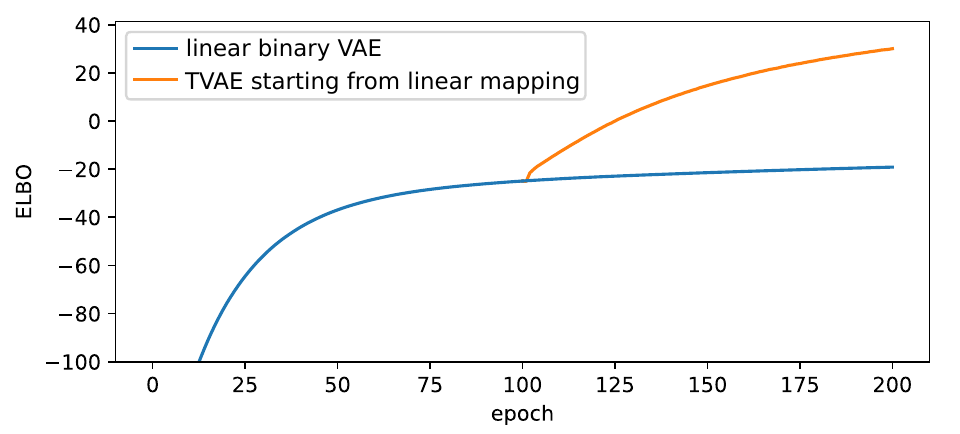
    \caption{ELBO gain of TVAE compared to linear VAE with binary latents (on $16\times{}16$ image patches).}
    \label{FigNaturalImages}
\end{figure}
\end{minipage}
\hfill
\begin{minipage}{0.5\linewidth}
\begin{figure}[H]
    \centering
    \def\svgwidth{\textwidth}
    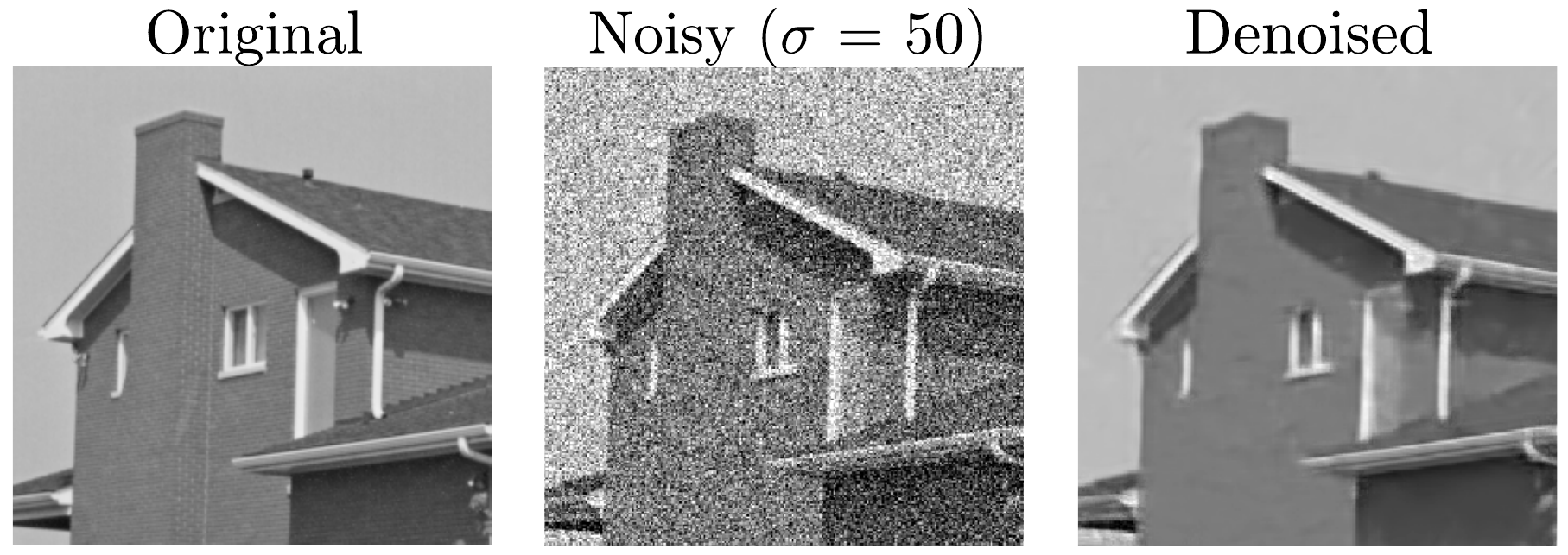
    \caption{TVAE denoising of house image with noise level $\sigma=50$. The denoised image has PSNR=30.03, the best of the runs of Tab.\,\ref{TabDenoising}.}
    \label{FigHouseDenoising}
\end{figure}
\end{minipage}

Compared to shallow linear models, we observed a similar efficiency and scalability of TVAE. The main additional computational costs
are given by passing the latent states through the a full decoder DNN instead of just
through a linear mapping. The sets of states used could be kept small, at size $S=|\PhiN|=64$, such
that $N\times{}(|\PhiN| + |\PhiNewN|)$ states had to be evaluated for each epoch. This compares to $N\times{}M$
states that would be used for standard VAE training (given $M$ samples are drawn per data point).
Differently to standard
VAE training the $\PhiN$ have to be remembered across iterations. For very large datasets, the additional $\mathcal{O}(N${}$\times${}$|\PhiN|${}$\times{}${}$H$) memory demand can be distributed over compute nodes, however.
To further investigate scalability, we went to %to VAEs with 
up to $H$=$1000$ latent variables (while using 100 units in the DNN middle layer). TVAE training time remained in line with the theoretical linear scaling with $H$ while the variational bound further increased.

%and the variational bound increased to about 84.9 nats.
%VAE training $\PhiN$ have to be remembered across iterations. For very large datasets, the additional $\mathcal{O}(N\times{}|\PhiN|\times{}H)$ memory demand %can be distributed over computing nodes, however.
%To further investigate scalability, we went to VAEs with up to $H$=1000 latent variables (while using 100 units in the DNN middle layer). TVAE training remains efficient as suggested by the theoretical linear scaling with $H$ and the ELBO increases to about 84.9 nats.

%
%While initializing TVAE weights with BSC can highlight the benefit of the DNN non-linearity, we observed this BSC pre-training to not be necessary in general.
%We also ran larger experiments in which we trained TVAE from scratch on the same dataset, scaling to $H$=1000 latent variables and 100 units in the network middle layer, with further increases in ELBO values up to a value of 84.9 nats.

\begin{table}
\begin{minipage}[t]{.3\textwidth}
   \setlength{\abovecaptionskip}{0pt}%
\caption{Denoising performance in PSNR (dB) for the `house' image under controlled conditions ($D$=8$\times{}$8, $H$=$64$ for all algorithms).}
\label{TabH64}
\end{minipage}\hfill
\begin{minipage}[t]{.7\textwidth}\vspace*{0pt}%
%\vspace{-2ex}}
\begin{center}
\begin{small}
\begin{sc}
\begin{tabular}{c c c c}
\toprule
% & \multicolumn{3}{c}{Noise level ($\sigma$)} \\
  & $\sigma$=15 & $\sigma$=25 & $\sigma$=50 \\
\midrule
MTMKL & \textbf{34.29} & 31.88 & 28.08 \\
GSC & 32.68 & 31.10 & 28.02 \\
%S5C & 33.50 & 32.08 & 28.35 \\
var-BSC & 32.25 & 31.15 & 28.62 \\
TVAE & 34.27 $\pm$ .02 & \textbf{32.65 $\pm$ .06}  & \textbf{29.61 $\pm$ .02}\\
\bottomrule
\end{tabular}
\end{sc}
\end{small}
\end{center}
\end{minipage}
\end{table}

\begin{table}
\begin{minipage}[t]{.3\textwidth}
   \setlength{\abovecaptionskip}{0pt}%
\caption{Denoising performance in PSNR (dB) for the `house' image for different algorithms with different optimized hyper-parameters. The \textbf{top} category only requires the noisy image. The \textbf{middle} requires additional information such as noise level (KSVD, WNNM, BM3D) or additional noisy images with matched noise level (n2v$^{\dagger}$). The \textbf{bottom} requires large clean datasets.}
\label{TabDenoising}
\end{minipage}
\hfill
\begin{minipage}[t]{.7\textwidth}\vspace*{0pt}%
\begin{center}
\begin{small}
\begin{sc}
    \begin{tabular}{c c c c}
        \toprule
%         & \multicolumn{3}{c}{Noise level ($\sigma$)} \\
         & $\sigma$=15 & $\sigma$=25 & $\sigma$=50 \\
        \midrule
        n2v$^{\star}$ & 32.05 & 29.20 & 25.42\\
        \small{MTMKL} & \textbf{34.29} & 31.88 & 28.08 \\
        GSC & 33.78 & 32.01 & 28.35 \\
        S5C & 33.50 & 32.08 & 28.35 \\
        var-BSC & 33.50 & 32.32 & 28.91 \\
        TVAE & 34.27 $\pm$ .02 & \textbf{32.65 $\pm$ .06} & \textbf{29.98 \small{$\pm$ .05}} \\
        \midrule
        n2v$^{\dagger}$ & 33.91 & 32.10 & 28.94 \\
        KSVD & 34.32 & 32.15 & 27.95 \\
        WNNM & 35.13 & 33.22 & 30.33 \\
        BM3D & 34.94 & 32.86 & 29.37 \\
        \midrule
        EPLL & 34.17 & 32.17 & 29.12 \\
        BDGAN & 34.57 & 33.28 & 30.61 \\
        DPDNN & \textbf{35.40} & \textbf{33.54} & \textbf{31.04} \\
\bottomrule
\end{tabular}
\end{sc}
\end{small}
\end{center}
\end{minipage}
\end{table}

{\bf Effectiveness: Image Denoising.} As we have observed, scaling to large latent spaces does not pose a problem for the presented approach. It is clear, however, that memory and computational cost increase with the number of data points. Above, we processed $100,000$ data points which is still feasible for the small DNNs used. However, larger DNNs increase computational load significantly because $N\times{}(|\PhiN| + |\PhiNewN|)$ latent states have to be passed through the decoder. Furthermore, larger DNNs require more data points to not overfit which further increases computational load of our $N$-dependent method. In many applications, there is, however,
anyway relatively few data available which makes the application of large DNNs prohibitive. One example is the task of
`zero-shot' denoising, i.e., denoising of an image when only the image itself is available. 
Learning without clean data recently became very popular. The task is currently addressed using 
approaches based on standard feed-forward DNNs whose training objectives have been altered to allow for training on noisy images \citep[e.g.][]{LehtinenEtAl2018,krull2019noise2void}. Deep generative models are, on the other hand, more naturally suited for training on noisy data as their learning objective can be used directly. \citet{ShocherEtAl2018} also argue that smaller DNNs are sufficient for the `zero-shot' setting. Because of its recent popularity and suitability for approaches with smaller DNNs, we consequently focus on `zero-shot' denoising. 
As a very significant additional benefit, the task allows for directly comparing the VAE approach to a large range of other approaches that have recently been suggested. Most notably we
can compare to non-deep generative models, large feed-forward DNNs \citep[][]{ZhuEtAl2019,DongEtAl2019} and DNNs dedicated to learning from noisy data \citep[][]{LehtinenEtAl2018,krull2019noise2void}.

%
%Having established scalability of TVAEs to large latent spaces, we now focus on the third question. The denoising benchmark that offers the broadest possible comparison to other methods is probably the `house' image (Fig.\,\ref{FigHouseDenoising} left). The standard benchmark settings for `house' make use of additive Gaussian white noise with standard deviations $\sigma\in\{15,25,50\}$.
%
%The most standard such benchmark is denoising the `house' image (Fig.\,\ref{FigHouseDenoising} left) corrupted by %additive Gaussian white noise with standard deviations $\sigma\in\{15,25,50\}$.
%
%Finally, we turn to the common and interesting real-world problem of denoising a single image for which no corresponding noiseless training dataset is available. This is the
%problem encountered, for example, when restoring old deteriorated photographs.
%To this end, we applied zero-mean Gaussian noise with standard deviation $\sigma=\{15,25,50\}$ to the well-known house image and 
%
%Given a noise value $\sigma$, we trained a TVAE on square patches extracted from the noisy image. We then used the trained model to estimate the most %likely value of each of the image pixels (see Appendix B).
%
%The image denoising benchmark used here also enables direct comparison to other variationally optimized generative models including MTMKL %\citep{TitsiasLazaro2011}, GSC \citep{SheikhEtAl2014} and S5C \citep{SheikhLucke2016}, which all showed SOTA performance for probabilistic sparse coding %at publication. 
%
The one denoising benchmark that offers the broadest possible comparison to other methods is probably the `house' image (Fig.\,\ref{FigHouseDenoising} left). The standard benchmark settings for `house' make use of additive Gaussian white noise with standard deviations $\sigma\in\{15,25,50\}$.
First, consider the comparison in Tab.\,\ref{TabH64} where all models used the same patch size of $D=8\times8$ pixels and $H=64$ latent variables (Appendix B for details). Tab.\,\ref{TabH64} lists the different approaches in terms of the standard measure of peak signal-to-noise ratio (PSNR). Values for MTMKL \citep{TitsiasLazaro2011}, GSC \citep{SheikhEtAl2014} and S5C \citep{SheikhLucke2016}
were taken from the respective original publications (which all established new state-of-the-art results when first published).
As can be observed, TVAE significantly improves performance for high noise levels.
TVAE is able to learn the best data representation for denoising and represents the state-of-the-art in this controlled setting (i.e., fixed $D$ and $H$).
%According to Tab.\,\ref{TabH64} two factors enable the good performance of TVAE compared to the other approaches. First, the evolutionary optimization
%training algorithm itself 
%seems to be beneficial as a comparison of BSC to MTMKL and GSC suggests (see Appendix B for details).
%
%Second, denoising performance of TVAE is significantly better than BSC (1dB for this establish benchmark represents a major improvement).
The decoder DNN of TVAE provides the decisive performance advantage: TVAE significantly improves performance compared to the linear Binary Sparse Coding \citep[var-BSC,][]{HennigesEtAl2010,SheltonEtAl2011}, confirming that the high lower bounds of TVAE on natural images translate into improved performance on a concrete benchmark.
For $\sigma=25$ and $\sigma=50$, TVAE also significantly improves on MTMKL, GSC, and S5C.
These three approaches are based on a spike-and-slab sparse coding model \citep[also compare][]{GoodfellowEtAl2012}. %While TVAE is using a less flexible Bernoulli prior, 
Despite the less flexible Bernoulli prior, the decoder DNN of TVAE provides the highest PSNR values for high noise levels.

In order to further extend our comparison, in the last experiment we considered the denoising task without controlling for equal conditions. Concretely, we allowed for any approach that performs denoising on the benchmark including approaches that are trained on large image datasets and/or use different patch sizes (including multi-scale and whole image processing). Note that
different approaches may employ very different sets of hyper-parameters that can be optimized for denoising performance: for sparse coding approaches, hyper-parameters include patch and dictionary sizes; for DNN approaches they include all network and training scheme hyper-parameters. By allowing for comparison in this less controlled setting, we can include a number of recent approaches including large DNNs trained on clean data and training schemes specifically targeted to noisy training data.
Tab.\,\ref{TabDenoising} shows the denoising performance for the three noise levels we investigated, with results for other algorithms taken from their corresponding original publications unless specified otherwise. For WNNM and EPLL we cite values from \citet{ZhangEtAl2017}. The results reported for noise2void \citep[n2v,][]{krull2019noise2void} were produced specifically for this work (see Appendix B).

Note that the best performing approaches in Tab.\,\ref{TabDenoising} cannot be trained on noisy data:
EPLL \citep{ZoranWeiss2011}, BDGAN \citep{ZhuEtAl2019} and DPDNN \citep{DongEtAl2019} all make use of clean training data (typically hundreds of thousands of data points or more).
For denoising, EPLL also requires the ground-truth noise level of the test image. Ground-truth noise level information is also required by KSVD \citep{EladAndAharon2006} and WNNM \citep{GuEtAl2014}.
As noisy data is very frequently occurring, removing the requirement of clean data has been of considerable interest
with, e.g., approaches like noise2noise \citep[n2n,][]{LehtinenEtAl2018} and noise2void being very actively discussed currently.
The n2n approach can achieve denoising performance on noisy training data
which is almost as high as the performance of a given DNN when trained on clean data. It would thus outperform all approaches
in Tab.\,\ref{TabDenoising} except for the bottom three. However, n2n requires different noise realizations of the very same underlying image.
noise2void aims to remove this artificial assumption.
Considering Tab.\,\ref{TabDenoising}, PSNR values of TVAE were consistently higher than those of n2v even if n2v was trained on external data with matched-noise level (n2v$^{\dagger}$ in Tab.\,\ref{TabDenoising}).
Performance of TVAE is 0.2dB lower than BM3D for $\sigma=25$ and 0.6dB higher for $\sigma=50$, which makes it,
for large noise levels, the state-of-the-art on this benchmark in the `zero-shot' setting (i.e., the setting n2n and n2v aim to address).

% in which only the noisy image is available \citep[`zero-shot' learning, see e.g.,][]{ShocherEtAl2018,ImamuraEtAl2019} or only noisy data (i.e., the setting n2n and n2v aim to address).

\section{Discussion}

%Here we investigated, for VAEs with binary latents, a direct optimization
%
We investigated a novel way to train VAEs with binary latents.
%
%Our approach is motivated by further complications of standard VAE training when discrete latents are used.
%The further complication of standard VAE training caused by discrete latents motivates investigations of alternative training approaches.
In order to avoid derivatives w.r.t.\ stochastic discrete latents, we here changed the standard training setup substantially.
Updates of the decoder DNN now involve a weighted sum over states (\ref{EqnFDeriW}) and the encoder DNN is replaced by
a discrete evolutionary optimization. % which replaces armortized inference.
The direct optimization of the encoder replaces methods that are usually considered indispensable for the training of VAEs: sampling approximation and reparameterization trick. Furthermore, the here investigated encoding model does not use a joint mapping for all datapoints to latent space,
i.e., the approach is not amortized. While amortization can be advantageous as information can be shared across datapoints, disadvantages in
terms of less tight lower bounds have also been pointed out \citep[e.g.][]{kim2018semi,cremer2018inference}. Related to this point, standard
VAE training usually involves factored Gaussian approximations of VAE posteriors which can introduce biases \citep[e.g., discussion by][]{VertesSahani2018}. The investigation of alternatives may therefore, more generally, shed further light on the consequences of specific approximation choices used to define VAE encoders. % \citep[also compare][]{Maneesh2018}.

The price we pay for not using amortization is efficiency: we optimize variational parameters for each data point which is, of course,
more costly. However, direct optimization can scale VAEs to large latent spaces if smaller DNNs are used. When the use of large DNNs
is anyway prohibitive because of limited data, the here studied approach can play out its effectiveness. For the recently popular task
of `zero-shot' denoising, we observed state-of-the-art results in a domain where VAEs have not been reported to be competitive before.
The competitive performance is presumably due to the approach not being subject to an amortization gap as well as not being based
on factored variational distributions. 

Our conclusion is consequently that direct discrete optimization can serve as an alternative for training discrete VAEs. 
In a sense, the approach can be considered as a brute-force optimization which is slower than conventional amortized 
training but more effective for scales at which it can be applied. To our knowledge, the approach is also the first training method
for VAEs that is not using sampling-based gradient estimates, and the first which makes VAEs competitive for `zero-shot' denoising.  

\newpage

\appendix

\section{Details of Encoder and Decoder Optimization}

\begin{figure}[ht]
    \centering
    \def\svgwidth{0.98\textwidth}
    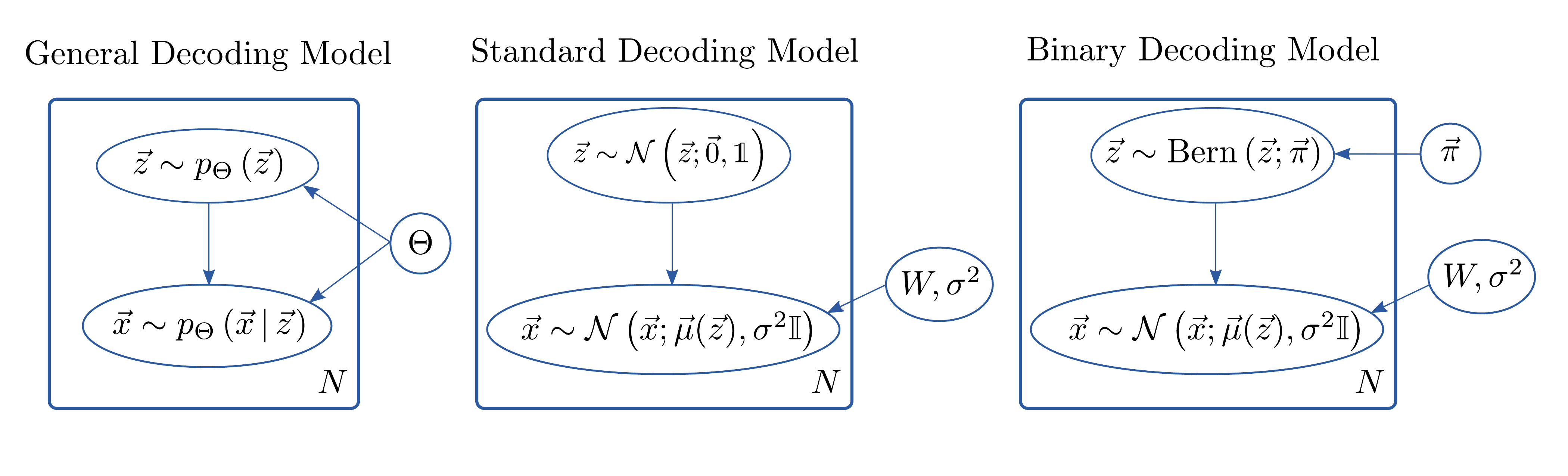
    \caption{From left to right: generic VAE decoding model, continuous-latent VAE model with Gaussian noise and the binary-latent VAE model of Eqn.\,(\ref{EqnBinaryVAE}), in plate notation.}
    \label{FigDecodingModels}
\end{figure}

\begin{figure*}[th]
\begin{center}
		\includegraphics[width=0.7\textwidth]{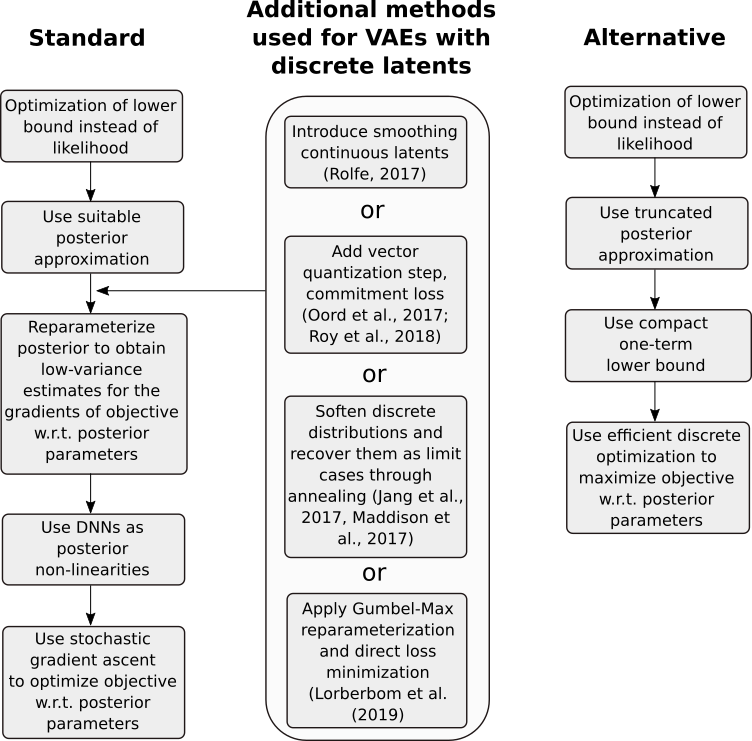}
\end{center}
\caption{Standard series of methods applied to optimize the encoding model of VAEs. \textbf{Left:} methods applied for encoding models of standard VAEs. \textbf{Middle:} additional methods applied to maintain the standard procedure of encoding model optimization also for discrete latent variables. \textbf{Right:} alternative approach to optimize the VAE encoding model using direct discrete optimization.}
\label{FigEncoding}
\end{figure*}

%We have described how the reformulation of the variational lower bound $\FF(\Phi,\Theta)$ in Eqn.\,\refp{EqnFMain} can give rise to
%a direct optimization of $\FF(\Phi,\Theta)$ w.r.t.\,$\Phi$ using truncated posteriors instead of the standard VAE encoding model. The parameters $\Theta$ of the generative model $\pTheta(\xVec)$ of Eqn.\,\refp{EqnBinaryVAE}, on the other hand, can still be optimized with standard techniques, namely gradient descent.
%
See Fig.\,\ref{FigDecodingModels} for a graphical comparison between the decoding models of a vanilla VAE and the binary VAE considered here (\ref{EqnBinaryVAE}).
Fig.\,\ref{FigEncoding} graphically illustrates different steps to optimize standard VAEs, and additional steps suggested by different contributions in order to optimize discrete VAEs. 

For the optimization of the binary VAE (\ref{EqnBinaryVAE}), consider the original form of the lower bound, Eqn.\,\refp{EqnFWithExp}.
When taking derivatives of $\FF(\Phi,\Theta)$ w.r.t.\,$\Theta$ we can ignore the entropy term\footnote{For our choice of variational distributions, it is not trivial that the entropy term actually can be ignored because the encoding model
$\qPhi(\zVec;\xVec)$ in (\ref{EqnQMain}) is defined in terms of the decoding model and its parameters.
For truncated distributions, however, it can be shown that the entropy term can still be ignored \citep{Lucke2019}.}.
For the binary VAE model of Eqn.\,(\ref{EqnBinaryVAE}) the gradient of the lower bound w.r.t.\,$W$ is then given by:
%
%\begin{align}
%\begin{split}
%%
%\vec{\nabla}_{\Theta}\FF(\Phi,\Theta) = &\sum_{n}  \EE{ \vec{\nabla}_{\Theta}\log\big( \pTheta(\xVecN\,|\,\zVec) \big) }_{\qPhiN}\\ 
%                                        &+ \sum_{n}  \EE{ \vec{\nabla}_{\Theta}\log\big( \pTheta(\zVec) \big) }_{\qPhiN}\,.
%\label{EqnFDeriOne}
%
%                                      &=& \phantom{+}\sum_{n}\frac{\sum_{\zVec\in\PhiN} \pTheta(\xVecN\,|\,\zVec)\,\pTheta(\zVec) \ \vec{\nabla}_{\Theta} \log\big( \pTheta(\xVecN\,|\,\zVec) \big)}
%                                                                  {\sum_{\zVec\in\PhiN} \pTheta(\xVecN\,|\,\zVec)\,\pTheta(\zVec)} \\
%
%                                       &&          + \sum_{n}\frac{\sum_{\zVec\in\PhiN}\pTheta(\xVecN\,|\,\zVec)\,\pTheta(\zVec) \ \vec{\nabla}_{\Theta} \log\big( \pTheta(\zVec) \big)}
%                                                                  {\sum_{\zVec\in\PhiN}\pTheta(\xVecN\,|\,\zVec)\,\pTheta(\zVec)}
%
%\end{split}
%\end{align}
%
%Plugging in the binary VAE model in (\ref{EqnBinaryVAE}) we obtain, for the parameters $W$ of the DNN $\vec{\mu}(\zVec,W)$:
%
\begin{eqnarray}
  \vec{\nabla}_{W}\FF(\Phi,\Theta) &=& \sum_{n} \vec{\nabla}_{W} \EE{  \log\big( \pTheta(\xVecN\,|\,\zVec)\,\pTheta(\zVec) \big) }_{\qPhiN}\nonumber\\
                           &=& \sum_{n} \vec{\nabla}_{W} \EE{\log\big( \pTheta(\xVecN\,|\,\zVec) \big)}_{\qPhiN}
                           \ =\  \sum_{n} \vec{\nabla}_{W} \EE{ \log\big( \NCal(\xVecN;\vec{\mu}(\zVec,W),\sigma^2\mathbb{I} \big) }_{\qPhiN} \nonumber\\
%
%                                  &=& \sum_{n} \vec{\nabla}_{W} \EE{\log\big( \pTheta(\xVecN\,|\,\zVec) \big)}_{\qPhiN}\nonumber\\
%
                                   &=& -\frac{1}{2\sigma^2} \sum_{n} \vec{\nabla}_{W} \sum\limits_{\ \zVec \in \PhiN} \qPhiN(\zVec)\phantom{i}\|\xVecN - \vec{\mu}(\zVec,W)\|^2 \nonumber\\
                                   &=& -\frac{1}{2\sigma^2} \sum_{n} \sum\limits_{\ \zVec \in \PhiN} \qPhiN(\zVec)\phantom{i}\vec{\nabla}_{W} \|\xVecN - \vec{\mu}(\zVec,W)\|^2\,,\label{EqnAppFDeriW}
\end{eqnarray}
where the weighting factors $\qPhiN(\zVec)$ are by using (\ref{EqnQMain}) and (\ref{EqnBinaryVAE}) given by:
\begin{eqnarray}
\qPhiN(\zVec) &=& \frac{\pTheta(\xVecN\,|\,\zVec)\,\pTheta(\zVec)}{\sum_{\zVecPrime\in\PhiN}\,\pTheta(\xVecN\,|\,\zVecPrime)\,\pTheta(\zVecPrime)} \nonumber\\
              &=& \frac{  \exp\big( -\frac{1}{2\sigma^2} \phantom{i}\|\xVecN - \vec{\mu}(\zVec,W)\|^2\,-\,\tilde{\vec{\pi}}^{\,T}\zVec \, \big)  }
                       { \sum\limits_{\ \zVecPrime \in \PhiN} \exp\big( -\frac{1}{2\sigma^2} \phantom{i}\|\xVecN - \vec{\mu}(\zVecPrime,W)\|^2\,-\,\tilde{\vec{\pi}}^{\,T}\zVecPrime \, \big)  } 
%
%\qPhiN(\zVec) := \phantom{\int}\pTheta(\zVec\,|\,\xVecN)\,\,\,\,\, \Big/ \,\, \sum_{\zVecPrime\in\PhiN}\,\pTheta(\zVecPrime\,|\,\xVecN)\,\ \mbox{\ if\ }\  \zVec\in\PhiN,
%                   &=\frac{\disS\phantom{\int}\pTheta(\xVecN\,|\,\zVec)\ \pTheta(\zVec)\phantom{\int}}{\hspace{-2mm}\disS\sum_{\zVecPrime\in\PhiN}\pTheta(\xVecN\,|\,\zVecPrime)\ \pTheta(\zVecPrime)}\,\delta(\zVec\in\PhiN),
%
\end{eqnarray}
for all $\zVec\in\PhiN$, where $\tilde{\pi}_h = \log\big(\frac{1-\pi_h}{\pi_h}\big)$. Note that the $\qPhiN(\zVec)$ are evaluated at the current
values of the parameters $\Theta$, they are therefore treated as constant, e.g., for the gradient w.r.t.\ $W$.

It may be interesting to compare the gradient estimate (\ref{EqnAppFDeriW}) to the gradient estimate of conventional VAE training. For this consider
a standard encoder given by an amortized variational distribution which we shall denote by $\qPhiNT(\zVec)$. The distribution $\qPhiNT(\zVec)$
could be a Gaussian whose mean and variance are set by passing data point $\xVecN$ through encoder DNNs. For discrete VAEs, $\qPhiNT(\zVec)$
can be thought of as an analog discrete distribution. If we now take gradients of (\ref{EqnFVAE}) w.r.t.\ $W$ and estimate using samples
from $\qPhiNT(\zVec)$, we obtain the familiar form:
\begin{eqnarray}
  \vec{\nabla}_{W}\FF(\Phi,\Theta) &=& \sum_{n} \vec{\nabla}_{W} \EE{  \log\big( \pTheta(\xVecN\,|\,\zVec)\,\pTheta(\zVec) \big) }_{\qPhiNT}\nonumber\\
                           &=&  \sum_{n} \vec{\nabla}_{W} \EE{ \log\big( \NCal(\xVecN;\vec{\mu}(\zVec,W),\sigma^2\mathbb{I} \big) }_{\qPhiNT} \nonumber\\
%
%                                  &=& \sum_{n} \vec{\nabla}_{W} \EE{\log\big( \pTheta(\xVecN\,|\,\zVec) \big)}_{\qPhiN}\nonumber\\
%
                                   &\approx& -\frac{1}{2\sigma^2} \sum_{n} \frac{1}{M} \sum_{m=1}^M \vec{\nabla}_{W} \|\xVecN - \vec{\mu}(\zVec^{(m)},W)\|^2\,,\ \ \mbox{\ where\ }\ \zVec^{(m)} \sim \qPhiNT(\zVec)\nonumber 
\end{eqnarray}
We can slightly rewrite this expression to obtain:
\begin{eqnarray}
  \vec{\nabla}_{W}\FF(\Phi,\Theta) &\approx& -\frac{1}{2\sigma^2} \sum_{n} \sum_{\zVec \sim \qPhiNT} \Big(\frac{1}{M}\Big)\, \vec{\nabla}_{W} \|\xVecN - \vec{\mu}(\zVec,W)\|^2\,,\label{EqnAppStandardGrad}
\end{eqnarray}
If we now compare with the gradient using the truncated approximation $\qPhiN(\zVec)$,
\begin{eqnarray}
  \vec{\nabla}_{W}\FF(\Phi,\Theta) &=& -\frac{1}{2\sigma^2} \sum_{n} \sum\limits_{\ \zVec \in \PhiN} \qPhiN(\zVec)\phantom{i}\vec{\nabla}_{W} \|\xVecN - \vec{\mu}(\zVec,W)\|^2\,,\label{EqnAppTruncGrad}
\end{eqnarray}
one can discuss analogous roles played by the subsets $\PhiN$ (the variational parameters of $\qPhiN(\zVec)$) and by a standard encoder $\qPhiNT$.
The states in a subset $\PhiN$ are used to estimate the gradient similar to the samples from a standard encoder $\qPhiNT(\zVec)$. The size of $\PhiN$
can consequently be thought of as analog to the number of samples used in a conventional estimation of the gradient.
Standard VAE training estimates the gradient by weighting all samples equally (with $(1/M)$) and the gradient direction is approximated using sufficiently many samples drawn from the current $\qPhiNT(\zVec)$. In contrast, truncated gradient estimation uses the states in $\PhiN$, and the gradient is computed using
a weighted summation with weights $\qPhiN(\zVec)$. These weights are computed by passing the states $\zVec$ through the {\em decoder} network.
The gradient is then, notably, not a stochastic estimation but exact: gradient ascent is guaranteed (for small steps) to always monotonically increase the variational lower bound.

{\bf Computational Complexity.} To add to the discussion of computational complexity of TVAE compared to standard VAE training, consider again
Eqns.\,\ref{EqnAppStandardGrad} and \ref{EqnAppTruncGrad}. If as many samples $M$ are used, per data point,
as there are states in each $\PhiN$, then both sums have the same number of summands. The evaluation of the
gradients of the mean square error (MSE) is consequently precisely the same for both approaches.
The additional weighting factors $\qPhiN(\zVec)$ have to be computed for TVAE. However, the weighting
factors just represent a small overhead because the evaluation of the
decoder DNN for the states in $\PhiN$ is a computation that can be reused from the
updates of $\PhiN$. 

The main computational differences are in the updates of $\PhiN$ compared to the update of encoder DNNs
for conventional VAEs. Once the parameters $\Theta=(W,\sigma^2,\piVec)$ are updated using (\ref{EqnAppTruncGrad}),
new states for $\PhiN$ have to be sought based on criterion (\ref{EqnCriterionVAE}). In practice and for each $n$,
we generate $M'$ new states according to the applied evolutionary procedure. To select the best states
we have to pass all these $M'$ new states through the decoder DNN to evaluate (\ref{EqnCriterionVAE}). Furthermore,
we have to pass all $M$ states already in $\PhiN$ through the DNN to re-evaluate (\ref{EqnCriterionVAE}) because
the parameters $\Theta$ have changed. In summary, we do require $\Ocal(N\times{}(M+M'))$ passes through
the decoder DNN. Selecting the $M$ best states from the $(M+M')$ states does not add complexity as this
can be done in $\Ocal(M+M')$ for each $n$ \citep[][]{BlumEtAl1973}. The EA does add to the computational load
but parent selection and mutation only add a constant offset for each of the considered states.

For comparison with standard VAEs, if we use $M$ samples of an encoder $\qPhiNT(\zVec)$, we require $\Ocal(M\times{}N)$ passes
through the decoder DNN to update the parameters $\Theta$ according to (\ref{EqnAppStandardGrad}). For the
encoder update, one requires $N\times{}\tilde{M}$ passes through encoder and decoder DNN to estimate the gradient
w.r.t.\ the encoder weights (if we draw $\tilde{M}$ samples for each data point from a conventional encoder distribution $\qPhiNT(\zVec)$.
The additional overhead to actually draw the samples is usually negligible.

Hence, the computational complexity of TVAE training is comparable if $M\approx{}M'\approx{}\tilde{M}$. However, conventional VAE training
is amortized, i.e., the update of encoder weights uses information from all data points $n$. In contrast, TVAE training is not amortized, i.e.,
the $\PhiN$ are updated per data point. The advantage of amortization is that in practice, weights of a conventional encoder can converge
faster or (alternatively) less samples $\tilde{M}$ are required. Considering the observed runtimes, more efficient conventional VAE training
can presumably in large parts attributed to faster convergence using amortization. Furthermore, the used number of samples $M$ for conventional VAE
training is usually smaller than best working sizes of $\PhiN$ (we used, e.g., $|\PhiN|=64$ and $|\PhiN|=200$ for denoising, see Tab.\,3); and
the required storage of $\PhiN$ results in overhead computations. On the other hand, amortization also has disadvantages \citep[e.g.][]{kim2018semi,cremer2018inference}.
The competitive performance for denoising may consequently be attributed at least in part to TVAE not being subject to an amortization gap. 
%
% Hence there is for the actual
%evaluation of (\ref{EqnAppTruncGrad}) just a small overhead for the prefactors as well as a requirement
%to evaluate new states $\zVecNew$. The more significant additional computation cost is consequently the evolutionary
%search for new states for $\PhiN$, which requires passing new states through the DNN to evaluate their fitness (\ref{EqnCriterionVAE}).
%TVAE executes this search for each data point. 
%
%In the case of a standard encoder, one firsts computes the parameters of the encoder distribution $\qPhiNT$ which requires
%one pass through the encoder DNN (or DNNs) for each data point $n$. Drawing samples from $\qPhiNT$ is then comparably fast
%(at least for standard distributions).
%
%drawing samples once mean and variances of the encoder m

%Drawing samples at least from elementary distributions for a standard encoders is
%comparably fast, as is the amortized optimization of standard encoder weights (although the latter efficiency
%is difficult to quantify). As amortization also has downsides, however, the higher computational cost of a non-amortized
%evolutionary search may not be perceived as very surprising.

\section{Details on the numerical experiments}

%\subsection*{Artificial Bars Data}
\subsection{Verification on Ground-Truth Data}

\begin{figure}[th!]
    \centering
    \def\svgwidth{0.35\textwidth}
    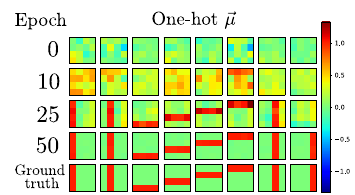
    \caption{TVAE training on simple bars data: noiseless output of the TVAE's DNN for the 8 possible one-hot input vectors over several training epochs. Generating parameters are in the last row.}
    \label{FigSimpleBars}
\end{figure}

\begin{figure}[th!]
    \centering
    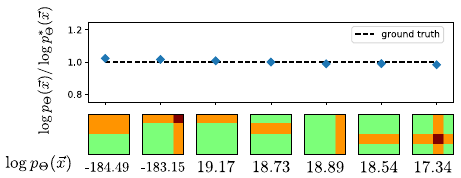
    \caption{Correlated bars test. The plot shows the ratio between inferred and ground-truth log-likelihoods $\log p_\Theta (\vec{x})$ of datapoints with interesting bar combinations. The inferred values are reported below the datapoints themselves.}
    \label{FigCorrelatedBars}
\end{figure}

We first evaluated TVAE training on artificial datasets with known
ground-truth parameters and log-likelihood, in order to verify the
correct functioning of the algorithm and to investigate possible 
local optima effects. The dataset consisted of 500 4x4 images generated by linear superposition of vertical and horizontal bars, with a small amount of Gaussian noise. The DNN's input and middle layers had 8 units each. The $\Phi^{(n)}$ variational sets consisted of 64 hidden states each. Fig.\,\ref{FigSimpleBars} shows the evolution of the run that achieved the highest ELBO value out of ten. All parameters were correctly recovered, and the ELBO value was consistent with actual ground-truth log-likelihood.

Such a simple test, however, can also be solved by linear models.
In order to demonstrate that TVAEs can solve non-linear problems, taking advantage of the neural network non-linearity embedded in the generative model, we introduced correlations between pairs of bars: the bars combinations shown in the first two datapoints from the left in Fig.\,\ref{FigCorrelatedBars} were discouraged from appearing together. We employed the same evolutionary scheme and again we selected the run with highest peak ELBO value out of ten. The model correctly learns that certain combinations of bars are much more unlikely than others, and correctly estimates their likelihood.

\begin{figure}[th!]
    \centering
    \def\svgwidth{0.9\textwidth}
    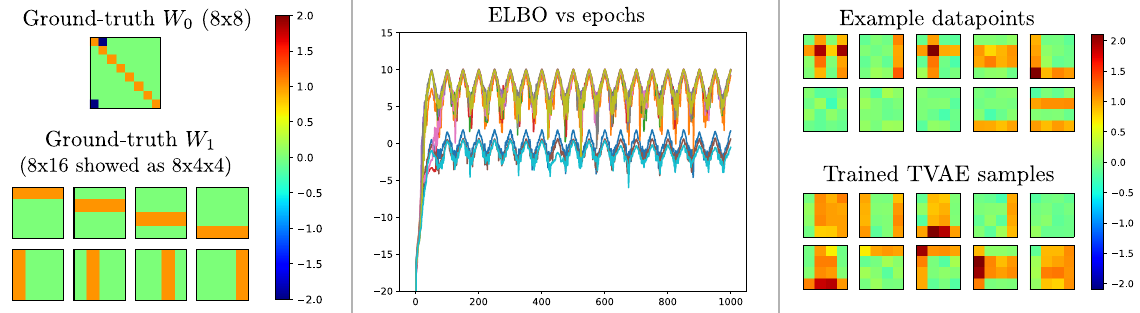
    \caption{From left to right: generative parameters for the correlated bars test; ELBO values over epochs for 10 runs; example datapoints and samples from the generative model.}
    \label{FigBarsTests}
\end{figure}

Fig.\,\ref{FigBarsTests} offers some more insight into the correlated bars test
experiment described. The left section of the figure shows the generative parameters for the dataset used: $W_0$ is the 8x8 weight matrix of the top-to-middle layer: this makes it so that the activation of the first latent variable inhibits activation of the second, and activation of the last latent variable inhibits activation of the last. Concretely, this results in a dataset where these specific bars combinations are discouraged from appearing. The weights $W_1$, visualized as 8 4x4 matrices, generate the actual bars. $\sigma^2$ was set to 0.01 and the dataset contained an average of two superimposing bars per datapoint ($\pi_h = 2/8$ for each $h$).

The middle section of the figure shows the ELBO values (averages over all batches for each epoch) as training progresses. The cyclic learning rate schedule is responsible for the oscillatory behavior.

The right section shows some example datapoints together with samples from the trained TVAE model that reached the highest ELBO value out of the ten runs.

\subsection{Denoising}
\label{AppDenoising}
%
% in each of the patches.
%Finally, we averaged \todo{what type of averaging?} the estimates of a given pixel over all patches in which the pixel appears to generate a final estimate for the non-noisy patch.
%Ideally, the model will capture not only the details of the noise distribution, but also recurring or interesting patterns of the underlying image.
%
Given a trained TVAE with parameters $\Theta$, we estimated the value of a pixel in a single patch as $x^{\text{est}}_d = \left< x_d \right>_{p_{\Theta}(x_d \mid \vec{x})}$. When
using $p_\Theta(x_d \mid \xVec) = \sum_{\{\zVec\}} p_\Theta(x_d \mid \zVec) p_\Theta (\zVec \mid \xVec)$ we obtain:
\begin{equation}
   x^{\text{est}}_d = \left< \left< x_d \right>_{p_\Theta(x_d \mid \vec{z})} \right>_{p_\Theta(\vec{z} \mid \vec{x})} = \left< \mu_d(\vec{z}) \right>_{p_\Theta(\vec{z} \mid \vec{x})} \,.
\label{EqnPixel}
\end{equation}
The expectation value on the right-hand-side of Eqn.\,\refp{EqnPixel} is then approximated based on the encoding parameters $\PhiN$ using truncated posteriors.
Finally, we took a weighted average of the estimates of a pixel value in different patches \citep[see, e.g.,][]{burger2012image} in order to generate the pixel values of the full denoised image.

In Tab.\,\ref{TabDenoisingParams} we list the exact hyper-parameters used to obtain the PSNR values reported. % in the paper in Tab.\,\ref{TabDenoisingParams}.
In parentheses, the parameters for the run on data with noise level $\sigma=50$ and unconstrained hyper-parameters are given, when they
differ from the other experiments.

\begin{table}[h!]
\begin{minipage}[t]{.3\textwidth}
\caption{Hyper-parameters for the denoising experiments on the house image.}
\begin{center}
\begin{small}
\begin{tabular}{l c}
\toprule
\multicolumn{2}{c}{\textbf{Neural network units}} \\
Input ($H$) & 64 (512) \\
Middle & 64 (512) \\
Output ($D$) & 64 (144) \\
\midrule
\multicolumn{2}{c}{\textbf{Cyclic Learning Rates}} \\
Min l.r.\ & 0.0001 \\
Max l.r.\ & 0.01 (0.05) \\
Epochs/cycle & 20 \\
Batch size & 32 \\
\midrule
\multicolumn{2}{c}{\textbf{Evolutionary parameters}} \\
Parents & 10 (5) \\
Children & 9 (4) \\
Generations & 4 (1) \\
Size of $\Phi^{(n)}$ & 200 (64) \\
\bottomrule
\label{TabDenoisingParams}
\end{tabular}
\end{small}
\end{center}
\end{minipage}
\hfill
\begin{minipage}[t]{.6\textwidth}\vspace*{0pt}%

\caption{Denoising performance of n2v in PSNR (dB) for the `house' image with AWG noise. For comparison, we additionally list the performance of TVAE (numbers copied from Tab.\,\ref{TabDenoising}). PSNR values for n2v$^{\star}$ are obtained by training only on the noisy image (i.e., in the same setting as used for MTMKL, GSC, var-BSC and TVAE in Tab.\,\ref{TabDenoising}. More training data improves performance for n2v. PSNR values for n2v$^{\dagger}$ show performance if additional training data in the form of noisy images
with AWG noise $\sigma=25$ is used. Further improvements (especially for high noise) are obtained if the n2v network is trained on training data with a noise level that matches
the noise of the test set (see n2v$^{\ddagger}$). For instance, we used for n2v$^{\ddagger}$ training data with $\sigma=50$ to denoise the `house' with $\sigma=50$. See text for further details.
%
%obtained by training on the noisy test image (no external training data used; see text for further details).
%
%
%PSNR values for n2v$^{\dagger}$ obtained by training on external data with $\sigma=25$ following the procedure described in the original n2v publication. Numbers for n2v$^{\star}$ %obtained similarly to n2v$^{\square}$ but with matched noise level. PSNR values for n2v$^{\dagger}$ obtained by training on the noisy test image (no external training data used; see %text for further details).
}
\label{Tabn2v}
\begin{center}
\begin{small}
\begin{tabular}{c c c c}
\toprule
% & \multicolumn{3}{c}{Noise level ($\sigma$)} \\
  & $\sigma$=15 & $\sigma$=25 & $\sigma$=50 \\
\midrule
n2v$^{\star}$ & 32.05 & 29.20 & 25.42\\
n2v$^{\dagger}$ & 32.93 & 32.10 & 20.96\\
n2v$^{\ddagger}$ & 33.91 & 32.10 & 28.94 \\
TVAE & \textbf{34.27 $\pm$ .02} & \textbf{32.65 $\pm$ .06} & \textbf{29.98 \small{$\pm$ .05}} \\
\bottomrule
\end{tabular}
\end{small}
\end{center}
\end{minipage}
\end{table}

%\subsection*{Discussion of Different Approaches}
%
To evaluate the performance on standard denoising benchmarks, we first compared TVAE to related probabilistic sparse coding approaches such as MTMKL, GSC and var-BSC (Tab.\,1). MTMKL and GSC use the data model of spike-and-slab sparse coding and for training mean-field and truncated posterior approximations with pre-selection are used, respectively. Compared to MTMKL and GSC, var-BSC uses a less complex data model and a training scheme also based on evolutionary optimization \citep[][]{GuiraudEtAl2018}.
%
%yet another training scheme, namely EEM which applies variational optimization of truncated posteriors. 
%
The denoising performance observed in the scenario with controlled conditions (Tab.\,1) shows that for high noise level ($\sigma=50$), var-BSC achieves higher PSNR values than MTMKL and GSC although the method uses a simpler data model. This observation demonstrates the effectiveness of the evolutionary training method used by var-BSC. However, PSNR values for TVAE are significantly higher due to the higher flexibility in modeling the data distribution provided by the used DNN.

In a second step, Tab.\,2 compared the performance of TVAE with respect to different denoising approaches including deterministic sparse coding (KSVD), a mixture model approach (EPLL), a non-local image processing method (WNNM) and state-of-the-art denoising methods based on deep neural networks (BDGAN and DPDNN). These approaches can be distinguished, e.g., by the amount of employed training data and by the requirement for clean data. 

TVAE as well as MTMKL, GSC and var-BSC do not require clean images for training. Furthermore, all these approaches can be trained if only the
single noisy image is available \citep[`zero-shot' learning; compare, e.g.,][]{ShocherEtAl2018,ImamuraEtAl2019}. Instead, EPLL, BDGAN and DPDNN use clean training data (typically tens or hundreds of thousands of data points are collected for training).
Approaches such as noise2noise \citep[n2n][]{LehtinenEtAl2018} and noise2void \citep[n2v][]{krull2019noise2void} occupy a middle ground: they can be trained on noisy data but they typically require much larger amounts of data than, e.g., TVAE or MTMKL. In the original n2v publication, for instance, 400 (noisy) $180\times180$ BSD \citep{BSD68} images were used to create a training dataset (this procedure also involved data augmentation; compare \citealt{krull2019noise2void}).
For our comparison with results of Tab.\,2, we used the standard, publicly available code for n2v together with the default training set ($\sigma=25$) employed in the original n2v publication. We then applied the trained n2v network to denoise the `house' image with $\sigma=25$. The resulting PSNR value was $32.10dB$ which is $0.76dB$ lower than the PSNR value for BM3D ($32.86dB$). The difference is consistent with an on average $0.88dB$ lower performance of n2v compared to BM3D on the BSD68 test set \citep[see][]{krull2019noise2void}. 
The same network can also be used to denoise an image with lower or higher noise level. The n2v network trained on $\sigma=25$ does, for instance, result in PSNR values
of $32.93dB$ for the `house' image with $\sigma=15$ and in $20.96dB$ for the `house' image with $\sigma=50$ (see $n2v^\dagger$ in Tab.\,4). Especially for high noise levels performance can be much improved, however, if the n2v network is trained using images with the same noise level as the test image. In order to do so, we followed the
procedure described in the n2v publication while adapting the noise level of $\sigma=15$ in one case and $\sigma=50$ for the other case. Trained on a dataset with
matched noise, we then denoised the `house' image with $\sigma=15$ in the one, and $\sigma=50$ in the other case (results listed as n2v$^{\ddagger}$ in Tab.\,\ref{Tabn2v}). 
The PSNR values obtained for `house' in this matched-noise-level scenario are much higher compared to the scenario with unmatched noise level (e.g., for $\sigma=50$ the PSNR improvement is approximately 8\,dB). The much lower performance for mismatched noise for n2v is in this respect consistent with observations for standard DNN denoising for which
training with the ground-truth noise level has been pointed out as important for performance \citep[][]{ChaudhuryAndRoy2017,ZhangEtAl2018}.

The n2v approach can avoid having to know the exact noise level, e.g., if it is trained on just the single noisy image. In a last experiment, we hence investigated this `zero-shot' denoising feature of n2v and applied the algorithm to denoise the `house' image while using the same noisy image for training that we seek to denoise (we took the publicly available code of n2v as an example and manually adjusted hyperparameters as follows: we set the "Percentage of pixel to manipulate per patch" to a value of 0.4, as "Number of training epochs" we used 400 and we set the "Number of parameter update steps per epoch" to 33). The obtained PSNR values are listed as n2v$^{\ast}$ in Tab.\,\ref{Tabn2v}. 

From Tab.\,\ref{Tabn2v} it can be observed, that for all considered training settings of n2v and all noise levels, 
PSNR values of TVAE are consistently higher than those of n2v even if n2v is trained on external data with matched-noise level.
Additional parameter tuning may improve performance of n2v$^{\ast}$ to a certain extent but PSNRs are in general much lower than n2v$^{\ddagger}$.
While we followed for n2v$^{\ddagger}$ the standard hyperparameter setting of the original paper/code publication of n2v \citep[][]{krull2019noise2void}, we cannot exclude further
improvements with parameter fine tuning for the `house' benchmark. However, we remark that the difference of n2v$^{\ddagger}$ and BM3D for the `house' benchmark
is on the very same range as the differences between n2v and BM3D as reported on the BSD data set in the original n2v publication. The stronger performing
BM3D is according to denoising performance the preferable comparison and as such included in Tab.\,2. In terms of efficiency, the n2v approach is in general (once trained) faster than BM3D as well as TVAE, however.

PSNR values of noise2noise (n2n) are usually very closely aligned with PSNR values achievable by feed-forward DNNs. More concretely, n2n uses, for instance, a RED30 network \citep[][]{MaoEtAl2016} which achieves 31.07\,dB PSNR on the BSD300 data set if trained on clean data. If directly trained on noisy data, RED30 achieves 31.06\,dB \citep[][]{LehtinenEtAl2018}. n2n is thus strongly performing in terms of PSNR.
The caveat of n2n compared to n2v is, however, that the noisy data n2n uses is rather artificial. The pairs of images n2n is trained on consist of two different
noise realization of the same underlying clean image. For real data, such a setting is only approximately occurring at most, which has motivated the n2v approach.

Like n2v, BDGAN and DPDNN are optimized for specific noise levels (specific standard deviations are used to generate the noisy training examples). EPLL is trained exclusively on clean image patches; for denoising, the algorithm requires the ground-truth noise level of the test image as input parameter. Ground-truth noise level information is also required by KSVD and WNNM. 

Like all approaches in the top category of Tab.\,2, TVAE does not require ground-truth noise level information, nor clean images, nor large amounts of training data.
For the `zero-shot' setting, TVAE is consequently the best performing system on the `house' benchmark. Such a high performance is notably
achieved using a basic DNN and relatively small patch sizes of $D=8\times{}8$ (for $\sigma=15$ and $\sigma=25$) or $D=12\times{}12$ (for $\sigma=50$).
All feed-forward DNNs for denoising use much larger patches (e.g., n2v use $64\times{}64$). That a competitive denoising performance can be achieved
for small patches, in general, argues in favor for VAE approaches to denoising. 
Indeed, TVAE even comes close to state-of-the-art approaches (BDGAN and DPDNN) that use very intricate DNN architectures and large amounts
of clean training data. We believe that such results underline the potential of the here investigated approach although the novelty of
the approach is the focus rather than extensive benchmarking. 

On the other hand, an important limitation of TVAE is its computational demand. For our experiments on the `house' image with
noise level $\sigma=50$ in Tab.\,2 we used $N=60025$ patches of $D=12\times{}12$ pixels,
which amounts to all possible non-overlapping square patches of that size that can be extracted from the image. For training and denoising we used a TVAE with $H=512$ latent variables,
sizes of $|\PhiN|=64$, and $512$ units in the DNN middle layer of the decoder. TVAE training required $49$ seconds per training epoch
when executing on a single NVIDIA Titan Xp GPU and $2.5$\,GB of GPU memory. We ran for $500$ epochs which required between seven
and eight hours on the single GPU. We did not observe significant changes in variational bound values or in denoising performance after 500 epochs in
any of the experiments we conducted for Tabs\,1 and 2. Runtime complexity increased linear with the number of data points $N$, with the dimensionality
of the data $D$, with the number of the latents $H$, and with the size of the DNN used. Runtimes also increased approximately proportional w.r.t.\ the size
of $\PhiN$. Empirically we observed a sublinear scaling with $|\PhiN|$ presumably because of significant overhead computations: for example, increasing from $|\PhiN|=64$ to $|\PhiN|=128$ (while keeping all other parameters as above) computational time increases from $49$ seconds per training epoch to $75$ seconds.

For noise levels $\sigma=15$ and $\sigma=25$ in Tab.\,2 we used smaller patch sizes ($D=8\times{}8$) and fewer stochastic latents ($H=64$) but larger $\PhiN$ (i.e., $|\PhiN|=200$). In general, if the patch size $D$ is increased, more structure has to be captured. This can be done either by increasing the size
of the stochastic latents $H$ or by using larger DNNs. Both, in turn, requires more training data in order to estimate the increased number of parameters.
In the current setup, the sizes of $D$ which are currently feasible are comparably small. The denoising performance based on small patches is, however, notably
very high. 

For comparison, n2v uses up to $D=64\times{}64$ and also all other feed-forward DNN approaches use significantly larger patch sizes than TVAE (and the other
approaches in category~1). Still, n2v can be trained efficiently on large patches requiring approximately 19 hours on a NVIDIA Tesla K80 GPU for training on approximately 3k noisy images of shape 180x180 and seconds for the denoising of one 256x256 image. The higher computational demand of TVAE is also the reason why averaging across databases with many images
(such as BSD68) or applications to large single images quickly becomes infeasible. As a novel approach, TVAE is, however, far from being fully optimized algorithmically compared to large feed-forward approaches, and there is certainly further potential to improve training efficiency.

While denoising is, in general, well suited for deep generative models, performance for standard image denoising is by far
not as common as such results for standard DNNs (which may also be related to efficiency aspects).
An exception is a recent GAN approach \citep[BDGAN;][]{ZhuEtAl2019}. VAEs are often evaluated using
binarized MNIST with approximate log-likelihoods for comparison; that benchmark, however, is not consistent with the Gaussian noise model used here and does
not allow a direct comparison with feed-forward DNNs which are the state-of-the-art. 

\section{Zero-shot inpainting}

\begin{figure*}[th]
\begin{center}
		\includegraphics[width=\textwidth]{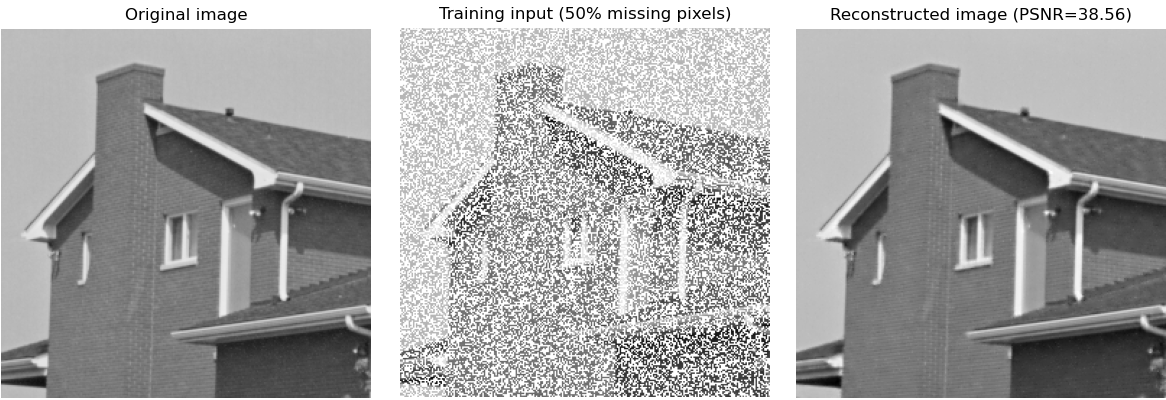}
\end{center}
\caption{Inpainting of the `house` image with TVAE.}
\label{FigInpainting}
\end{figure*}

The TVAE approach can be applied to a ‘zero-shot’ inpainting task using a procedure similar to the
one presented in the main text for denoising (also see \ref{AppDenoising}): a single image with
missing pixels (Fig.\,\ref{FigInpainting}, center) is divided into square patches to form the training set; during
training, missing pixels can be treated as unknown observables when evaluating log-joint probabilities of a data-point; Eq.\,\ref{EqnPixel}.
is then used to estimate likely values of the missing pixels, providing ``inpainted'' datapoints that can
be used for DNN backpropagation as usual. As TVAE is a non-amortized approach, missing values can directly be treated in a grounded probabilistic way. Amortized approaches will have to specify how an encoder network should treat missing values.

When evaluating TVAE on standard inpainting benchmarks, we observe competitive performance compared to other approaches. Tab.\,\ref{TabInpainting} shows a comparison of inpainting performance (in terms of PSNR) with previous state-of-the-art systems that like TVAE do not require large, clean training data nor information, e.g., on the noise level. As can be observed, TVAE outperforms approaches such as BPFA \citep{ZhouEtAl2012} or \citet{Papyan2017}. TVAE performance is lower than for DIP \citep{UlyanovEtAl2018}. TVAE, like BPFA and \citet{Papyan2017}, is a permutation-invariant approach, however. That is, the TVAE model is itself not using information about the 2D nature of images. DIP results rely, on the other hand, on a large dedicated DNN with LeakyReLU as activation functions, a U-net / hourglass architecture with skip connections, and convolutional units with reflection padding (see supplement of \citet{UlyanovEtAl2018}). The convolutional stages do explicitly assume the 2D image structure. We also remark that DIP uses in total 2 million parameters (and many more hyperparameters) compared to about 0.5 million parameters of the standard multi-layer perceptron used in TVAE.

Tab.\,\ref{TabParams} provides a list of the hyper-parameters used for these experiments.

\begin{table}[h!]
\caption{Inpainting performance in PSNR (dB) for the `house' image with 50\% of missing pixels.}
\begin{center}
\begin{tabular}[l c]{l c}
   \toprule
   Papyan et al.&34.58\\
   BPFA&38.02\\
   TVAE&38.56\\
   DIP&{\bf39.16}\\
   \bottomrule
\end{tabular}
\label{TabInpainting}
\end{center}
\end{table}

\begin{table}[h!]
\caption{Hyper-parameters used to produce the results of Fig.\,\ref{FigInpainting}. Training lasted 500 epochs taking around 60 seconds/epoch on a single NVIDIA Titan Xp GPU.}
\begin{center}
\begin{small}
\begin{tabular}{l c}
\toprule
\multicolumn{2}{c}{\textbf{Neural network units}} \\
Input ($H$) & 512 \\
Middle & 512 \\
Output ($D$) & 144 \\
\midrule
\multicolumn{2}{c}{\textbf{Cyclic Learning Rates}} \\
Min l.r.\ & 0.0001 \\
Max l.r.\ & 0.01 \\
Epochs/cycle & 20 \\
Batch size & 32 \\
\midrule
\multicolumn{2}{c}{\textbf{Evolutionary parameters}} \\
Parents & 5 \\
Children & 4 \\
Generations & 1 \\
Size of $\Phi^{(n)}$ & 64 \\
\bottomrule
\label{TabParams}
\end{tabular}
\end{small}
\end{center}
\end{table}

\end{document}